%% file: arxiv_main.tex
\documentclass[10pt,twocolumn,letterpaper,table]{article}

\usepackage[pagenumbers]{cvpr} % To force page numbers, e.g. for an arXiv version
\definecolor{cvprblue}{rgb}{0.21,0.49,0.74}
\usepackage[pagebackref,breaklinks,colorlinks,allcolors=cvprblue]{hyperref}
\usepackage{bm}
\usepackage[capitalize]{cleveref}
\usepackage{ulem}
\usepackage{xcolor}
\usepackage{url}
\usepackage{multirow}
\usepackage{wrapfig,lipsum,booktabs}
\usepackage{listings}
\usepackage{algorithm}
\usepackage{algpseudocode}
\usepackage{amssymb}
\usepackage{tocloft}  
\usepackage{setspace}

\def\ours{\textbf{See3D}\xspace}
\def\data{WebVi3D\xspace}

\title{\includegraphics[width=0.06\textwidth]{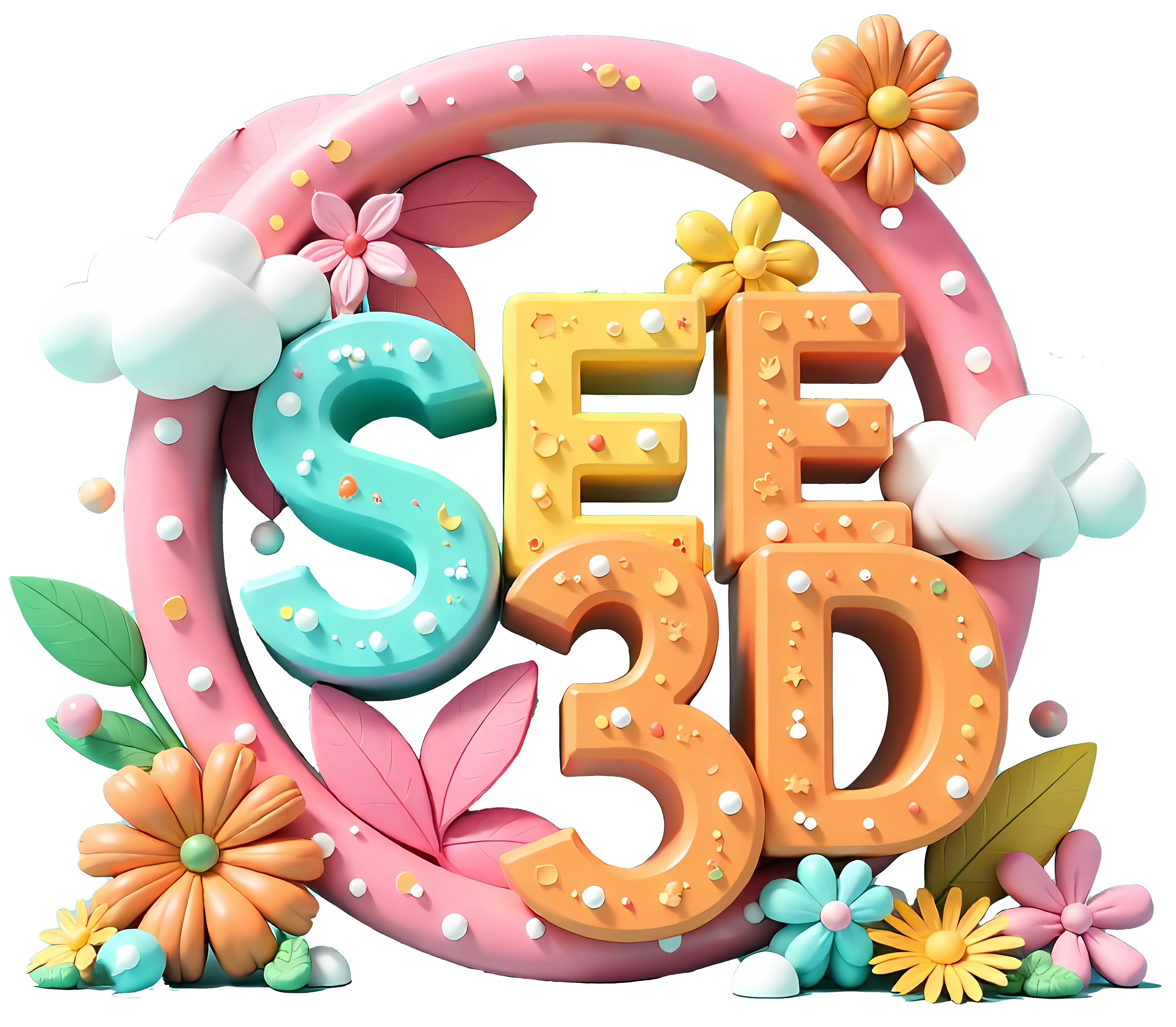} You See it, You Got it: Learning 3D Creation on Pose-Free Videos at Scale}

\author{ 
Baorui Ma\thanks{Equal contribution. ~\textsuperscript{\dag} Correspondence to XW and LT.},
~Huachen Gao$^{*}$, 
~Haoge Deng$^{*}$,
~Zhengxiong Luo, 
~Tiejun Huang,  
~Lulu Tang$^{\dagger}$,
~Xinlong Wang$^{\dagger}$    \\ 
~~~~~~~~~~~~~~~~~~~~~~~~~~~~~~~~~~~\\
Beijing Academy of Artificial Intelligence (BAAI)\\
}

\begin{document}
\input {arxiv/demo.tex}
\maketitle
\input{arxiv/0_abstract}

\input{arxiv/1_intro}

\input{arxiv/2_related_work}

\input{arxiv/3_method}

\input{arxiv/4_exp}

\input{arxiv/5_conclusion}

{
    \small
    \bibliographystyle{ieeenat_fullname}
    \bibliography{main}
}

\input{arxiv/6_arxiv_suppl}  

\end{document}

%% file: arxiv/0_abstract.tex
\begin{abstract}

Recent 3D generation models typically rely on limited-scale 3D `gold-labels' or  2D diffusion priors for 3D content creation.  However, their performance is upper-bounded by constrained 3D priors due to the lack of scalable learning paradigms.   
In this work, we present \textbf{\ours}, a visual-conditional multi-view diffusion model trained on large-scale Internet videos for open-world 3D creation.
The model aims to \textbf{Get} \textbf{3D} knowledge by solely \textbf{See}ing the visual contents from the vast and rapidly growing video data --- You See it, You Got it.
To achieve this, we first scale up the training data using a proposed data curation pipeline that automatically filters out multi-view inconsistencies and insufficient observations from source videos. This results in a high-quality, richly diverse, large-scale dataset of multi-view images, termed WebVi3D, containing 320M frames from 16M video clips.
Nevertheless, learning generic 3D priors from videos without explicit 3D geometry or camera pose annotations is nontrivial, and annotating poses for web-scale videos is prohibitively expensive.
To eliminate the need for pose conditions, we introduce  an innovative  \textit{visual-condition} - a purely 2D-inductive visual signal generated  by adding time-dependent noise to the masked video data.
Finally, we introduce a novel visual-conditional 3D generation framework by integrating \ours into a warping-based pipeline for high-fidelity 3D generation. 
Our numerical and visual comparisons on single and sparse reconstruction benchmarks show that \ours, trained on cost-effective and scalable video data, achieves notable zero-shot and open-world generation capabilities, markedly outperforming models trained on costly and constrained 3D datasets. Additionally, our model naturally supports other image-conditioned 3D creation tasks, such as 3D editing, without further fine-tuning. Please refer to our project page at: {\href{https://vision.baai.ac.cn/see3d}{https://vision.baai.ac.cn/see3d}.}

\end{abstract}

%% file: arxiv/1_intro.tex
\vspace{-8pt}

\section{Introduction}
\label{sec:intro}
Recent advances in 3D generation are essential for fields like virtual reality, entertainment, and simulation, offering the potential not only to recreate intricate real-world structures but also to expand human imagination.
Nevertheless, developing these models is constrained by the scarcity and high costs of accessible 3D datasets. Despite recent industry efforts \cite{tochilkin2024triposr, zhang2024clay, yang2024hunyuan3d}  create extensive proprietary 3D assets, these initiatives come with substantial financial and operational burdens. Currently, building  such a large-scale 3D dataset for academia remains prohibitively expensive. This motivates us to pursue a scalable, accessible, and affordable data source that can compete with advanced closed-source solutions, thereby enabling the broader research community to train high-performance 3D generation models.

\vspace{4pt}
Human perception of the 3D world does not rely on specific 3D representation (e.g., point clouds\cite{point}, voxel grids \cite{voxel}, meshes \cite{mesh}, or neural fields \cite{mildenhall2021nerf})  or precise camera conditions. Instead, our 3D awareness is shaped by multi-view observations accumulated throughout our lives. This raises the question:  \textit{Can models similarly learn universal 3D priors from large collections of multi-view images?}  Fortunately, Internet videos offer a rich source of multi-view images, captured from various locations with diverse sensors and complex camera trajectories, providing a scalable, accessible, and cost-effective data source. Thus, \textit{how can we effectively learn 3D knowledge from Internet videos?}

\vspace{4pt}
The core challenges in achieving this goal are twofold: 1) filtering relevant, 3D-aware video data from raw sources, specifically static scenes with varied camera viewpoints that provide sufficient multi-view observations; and 2) learning generic 3D priors from videos lacking explicit 3D geometry and camera pose annotations (i.e. pose-free videos). This work carefully addresses these challenges and  introduces a pose-free, visual-conditional multi-view diffusion (MVD) model, \ours, for open-world 3D creation.

\vspace{4pt}
Specifically, we establish a novel video data curation pipeline that automatically filters out  data with dynamic content or restricted camera viewpoints from source videos.
The resulting dataset, termed WebVi3D, comprises 15.99M video clips from 25.48M source videos, totaling 4.41 years in duration—orders of magnitude larger than previous 3D datasets, such as DLV3D (0.01M) \cite{ling2024dl3dv}, RealEstate10K (0.08M) \cite{real10k}, MVImgNet (0.22M) \cite{yu2023mvimgnet} and Objaverse (0.8M) \cite{deitke2023objaverse}.

\vspace{4pt}
MVD models have recently gained  widespread attention due to their advantages of integrating the generative capabilities of 2D diffusion models while maintaining consistency across multiple views \cite{shi2023mvdream,liu2023syncdreamer,liu2024reconx,sun2024dimensionx,zhao2024genxd}.
Typically, these models rely on precise camera poses \cite{shi2023mvdream,gao2024cat3d,hong2023lrm,liu2023zero, muller2024multidiff,bahmani2024vd3d,xu2024cavia,xu2024camco,he2024cameractrl,zhang2024recapture,li2024vivid} or warped images rendered according to camera position \cite{yu2024viewcrafter,tung2024megascenes} as conditional inputs to guide 3D-consistent view generation. We refer to these conditions, derived from pose or 3D annotations, as 3D-inductive conditions.  
However, annotating web-scale videos is prohibitively costly, or even intractable in some  cases, posing significant challenges for scaling. 
To address this, we propose a novel, pose-free \textit{visual-condition} derived from pixel-space hints within videos.  It is a purely 2D-inductive visual signal, created by introducing \textit{time-dependent noise} to masked input videos, free from any 3D-inductive bias. This enables training  MVD model at scale, without requiring pose annotations.

\vspace{4pt}
Intuitively, the proposed \textit{ visual-condition} can generalize effectively  to tasks that rely on pixel-space hints distinct from those in videos, such as warping-based 3D generation \cite{chung2023luciddreamer,shriram2024realmdreamer} and mask-based 3D editing \cite{chen2024gaussianeditor}, without requiring additional training, see \cref{fig:demo}.  For instance, in warping-based 3D generation, pixels from a reference image are rearranged through warping operations, creating a \textit{specific} visual-condition to indicate camera movement. However, these warped images often exhibit artifacts or distortions, causing a significant domain gap compared to video frames. Whereas, our \textit{visual-condition} functions as a \textit{generic} one, capable of accommodating such unnatural images.

\vspace{4pt}
To further harness the potential  of \ours, we introduce an innovative visual-conditional 3D generation framework utilizing  a warping-based pipeline. This framework first constructs the \textit{visual-condition} using  \ours, then iteratively refines the geometry of novel views to build comprehensive scene observations. Finally, the generated images are used for Gaussian Splatting reconstruction \cite{3dgs,huang20242d}, which can be rendered from arbitrary viewpoints or converted into meshes through post-processing \cite{lorensen1998marching}.
In summary, our key contributions are as follows.

\begin{itemize}

\item We present \ours, a scalable visual-conditional MVD model for open-world 3D creation, which can be trained on web-scale video collections without pose annotations.

\item We curate WebVi3D, a multi-view images dataset containing static scenes with sufficient multi-view observations, and establish an automated pipeline for video data curation to train the MVD model. 

\item We introduce a novel warping-based 3D generation framework with \ours, which supports long-sequence generation with complex camera trajectories. 

\item  We achieve state-of-the-art results in single and sparse views reconstruction, demonstrating remarkable zero-shot and open-world generation capability, offering a novel perspective on scalable 3D generation.
\end{itemize}

\begin{figure*}[tb]
\centering
\includegraphics[width=
0.95\linewidth]{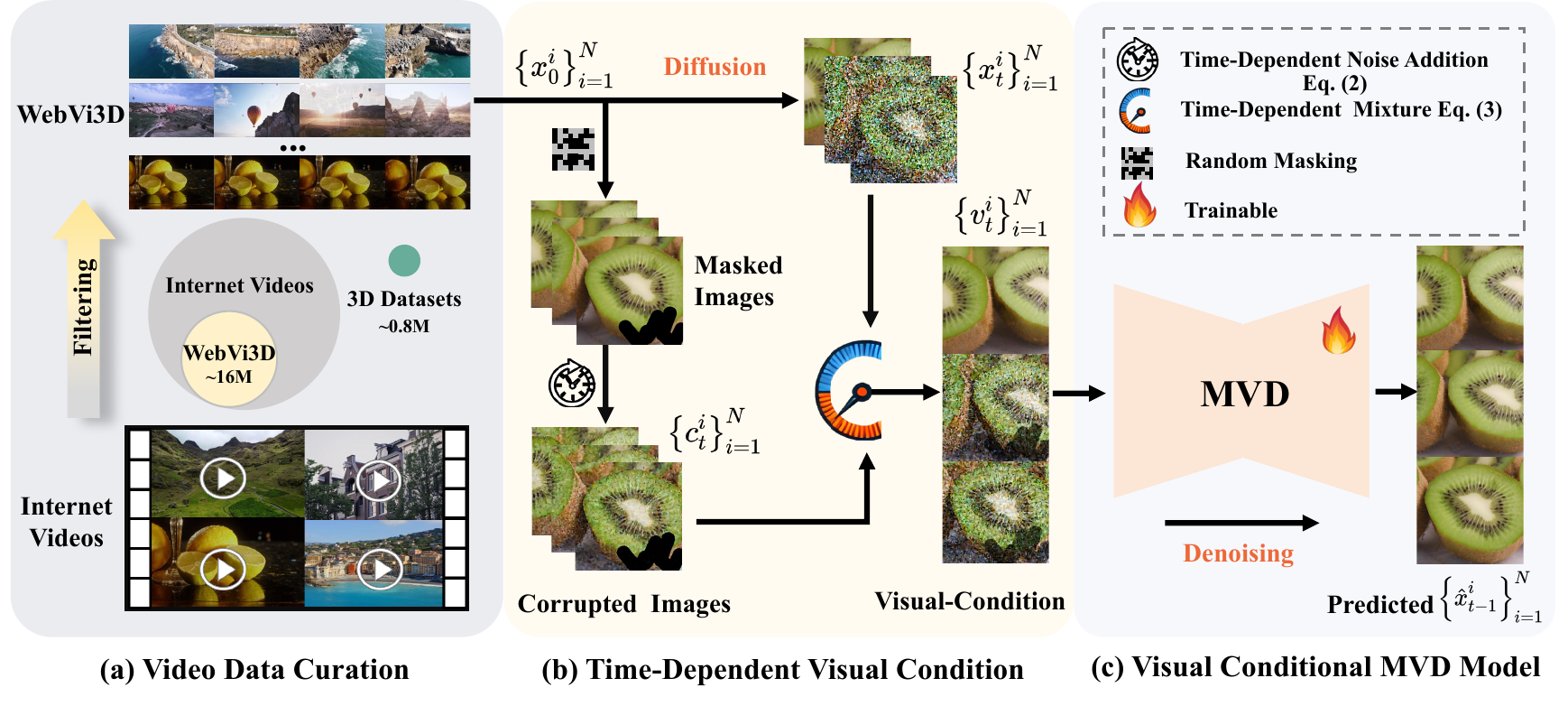}
 \vspace{-2pt}
\caption{\small \textbf{Overview of \ours.}
(a) We propose a four-step data curation pipeline to select multi-view images from Internet videos, forming the WebVi3D dataset, which includes $\sim$16M video clips across  diverse categories and concepts. 
(b) Given multiple views, we corrupt the original data into corrupted images $c_{t}^{i}$
at timestep $t$ by applying random masks and time-dependent noise. We then reweight the guidance of  $c_{t}^{i}$ and the noisy latent $x_{t}^{i}$ for the diffusion model to form  \textit{ visual-condition} $v_{t}^{i}$ through a time-dependent mixture. 
(c) MVD model is capable of training at scale to generate multi-view images conditioned on $v_{t}^{i}$, without requiring pose annotations. Since $v_{t}^{i}$ is a task-agnostic visual signal formed through time-dependent noise  and mixture,  it enables the trained model to robustly adapt to various  downstream tasks.}
\vspace{-2pt}
\label{fig:trainpipe}
\end{figure*}

%% file: arxiv/2_related_work.tex
\section{Related work}
\label{sec:related_work}

\noindent \textbf{Lifting 2D Generation into 3D.} Recent advances in 3D generation have been largely driven by the success of 2D diffusion models \cite{sohl2015deep,song2020denoising,ho2020denoising,rombach2022high}, which have revolutionized image and video generation. These works typically optimize 3D representations by maximizing the likelihood evaluated by 2D diffusion priors \cite{poole2022dreamfusion,latentnerf,tang2023dreamgaussian,wang2024prolificdreamer,kim2023collaborative,lin2023magic3d, yi2023gaussiandreamer, sun2023dreamcraft3d,12345}. 
An alternative approach uses  a warping-inpainting pipeline, integrating an offline depth estimator with a 2D diffusion-based inpainting model to iteratively generate 3D content \cite{chung2023luciddreamer, engstler2024invisible, yu2024wonderworld,yu2024viewcrafter,muller2024multidiff,hollein2023text2room,wang2024vistadream}.
However, 2D priors do not readily translate into coherent 3D representations. As a result, 2D lifting-based approaches often struggle to preserve high geometric fidelity, leading to issues like multi-view inconsistency and poor global geometry \cite{yu2024wonderjourney}. 

\vspace{7pt}
\noindent \textbf{Directly Learning 3D Priors.}
To better preserve geometric features, some works focus on directly learning 3D priors. For instance, feed-forward approaches \cite{hong2023lrm,gupta20233dgen, pflrm,mvlrm,wei2024meshlrm,longlrm,tochilkin2024triposr,tang2025lgm,xu2024instantmesh,xu2024grm,shen2024gamba,lu2024large,triplanegaussian,splatter,liu2025mvsgaussian,chen2025mvsplat,charatan2024pixelsplat,instant3d} take single/few views as input and directly output 3D representations using an encoder-decoder architecture, eliminating the need for additional optimization process per instance. Another line of research involves training diffusion models to predict 3D representations, such as point clouds \cite{lcm,pointe}, mesh \cite{lyu2023controllable,alliegro2023polydiff,shapee}, and implicit neural representation \cite{ma2024shapesplat, ssdnerf,wu2024direct3d,zhang2024clay}. 
However, these methods generally focus on object-level generation \cite{deitke2023objaverse,wu2023omniobject3d,tang2025lgm,triplanegaussian,zhang2024clay}, limiting their applicability to scene-level generation. Although recent research has made strides in building scene-level 3D datasets  \cite{knapitsch2017tanks,ling2024dl3dv,barron2022mipnerf360,dai2017scannet}, their scale remains relatively limited. The reliance on costly, limited-scale 3D datasets restricts  generalization to open-world or highly imaginative scenarios. 
In contrast, our approach  curates a large-scale, richly diverse dataset of multi-view images from Internet videos. By training the model at scale, it effectively supports both object-level and scene-level 3D creation. 

\vspace{7pt}
\noindent \textbf{Learning Multi-view Priors for 3D Generation. }
MVD  model inherits  the generative capabilities of 2D diffusion models while capturing multi-view correlations, achieving both generalizability and 3D consistency.  These merits have made it a focal point in recent 3D generation research \cite{shi2023mvdream,shi2023zero123++,12345++,yu2024viewcrafter, wang2023imagedream,long2024wonder3d,liu2023syncdreamer,han2025vfusion3d,qiu2024richdreamer,sargent2023zeronvs,gao2024cat3d}. 
However, as 2D diffusion models are typically  trained on 2D datasets, they lack precise control over image pose.
To address this, MVD-based approaches often  train their models on images paired with camera poses \cite{watson2022novel,liu2023zero,sargent2023zeronvs, wu2024reconfusion,gu2023nerfdiff}, where poses serve as  essential conditional inputs, represented by camera extrinsics \cite{sargent2023zeronvs,shi2023mvdream}, relative poses \cite{liu2023zero,shi2023zero123++,liu2023syncdreamer}, or Plücker rays \cite{gao2024cat3d,xu2024camco}. Yet, pose-conditional models  rely heavily on costly pose-annotated data, restricting training to smaller 3D datasets, thereby constraining their adaptability to out-of-distribution scenarios.
In contrast, we introduce a  novel visual-conditional approach   that supports scalable, pose-free MVD model training for open-world 3D generation.

%% file: arxiv/3_method.tex
\section{Method}
The primary objective of this work is to build a robust 3D generative model from the perspective of dataset scaling-up.
Previous works \cite{deitke2023objaverse, tung2024megascenes,reizenstein2021common} laboriously collect 3D data from designed artists, stereo matching, or Structure from Motion (SfM), which can be costly and sometimes infeasible.   In contrast, multi-view images offer a highly scalable alternative, as they can be automatically extracted from the vast and rapidly growing Internet videos. By using multi-view prediction as a pretext task, we demonstrate that learned 3D priors enable various 3D creation applications, including single view generation, sparse views reconstruction, and 3D editing in open-world scenarios.

\vspace{4pt}
The following sections outline our approach (Fig.\ref{fig:trainpipe}). \cref{sec:data} details the data curation pipeline, which selects static scenes with sufficient multi-view observations from raw video footage. \cref{sec:model} introduces our visual-conditional multi-view diffusion model, which effectively learns general 3D priors from pose-free videos. Finally, \cref{sec:3DGen} demonstrates a new visual-conditional 3D generation framework utilizing  a warping-based pipeline.

\subsection{Video Data Curation} \label{sec:data}
\begin{figure}[ht]
\centering
\includegraphics[width=\linewidth]{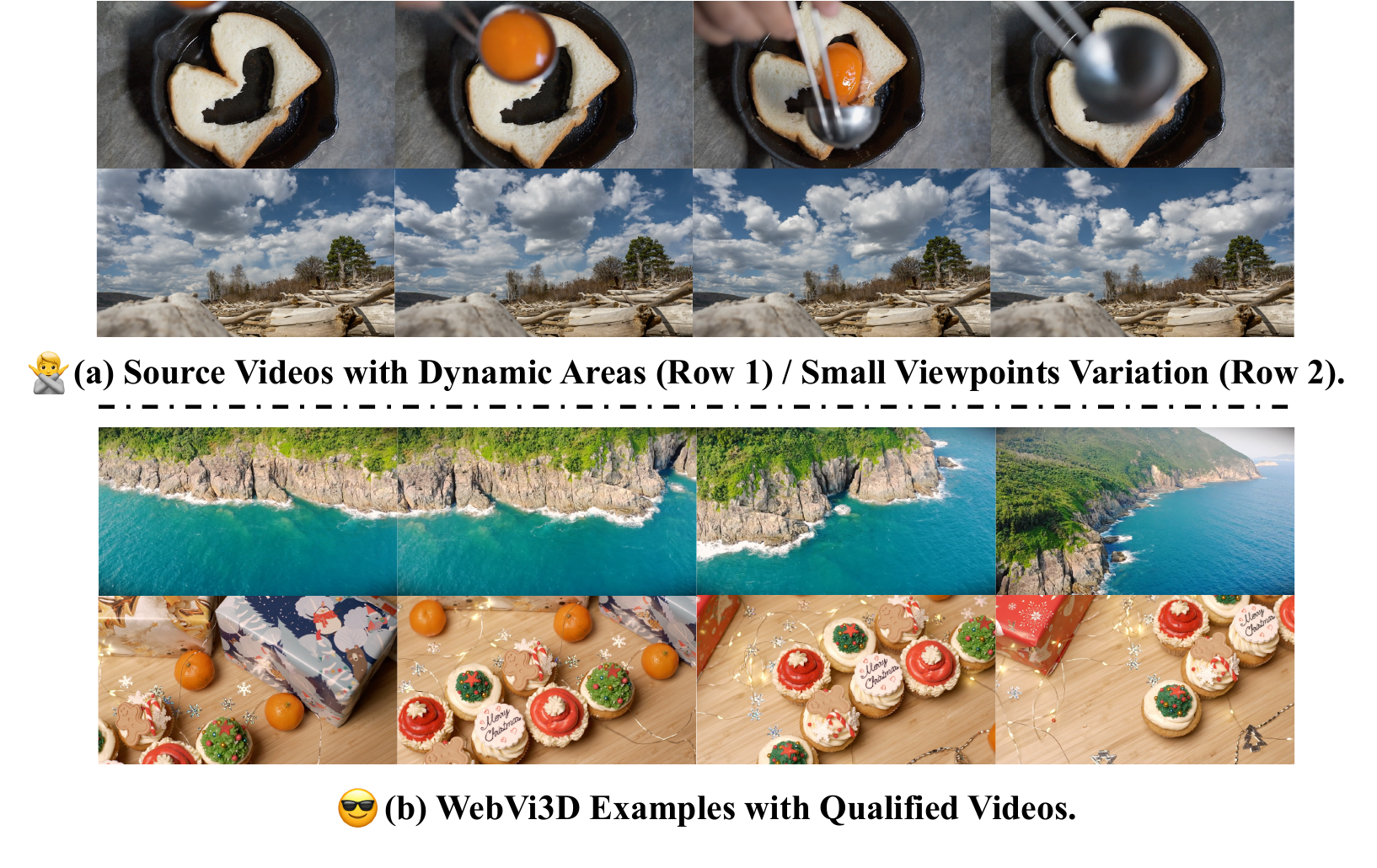}
 \vspace{-17pt}
\caption{\small(a-Row1): Dynamic content modifies  scene geometry across views; (a-Row2): Limited camera movement provides insufficient multi-view observations; (b) Our \data comprises static scenes with diverse camera trajectories.} 
\vspace{-12pt}
\label{fig:data}
\end{figure}

High-quality, large-scale video data rich in 3D knowledge is essential  for learning  accurate and reliable 3D priors. 
A well-defined 3D-aware video clip should exhibit two key properties: \textbf{1) temporally static scene content } and \textbf{ 2) significant  viewpoint variation} caused by the camera's ego-motion.
The first property ensures consistent 3D geometry across different viewpoints, while dynamic content can distort scene geometry and introduce biases that may degrade generation performance (\cref{fig:data}a-Row1). The second property guarantees  sufficient 3D observations from diverse  viewpoints. When the model is trained on videos with limited viewpoint variation (\cref{fig:data}a-Row2), it risks focusing on views adjacent to the reference view, rather than developing a comprehensive 3D understanding.

\vspace{4pt}
To obtain a massive volume of 3D data, we collect approximately  \textbf{25.48M} open-sourced raw videos, totaling \textbf{44.98 years} from the Internet, covering a wide range of categories, such as landscapes, drones, animals, plants, games, and actions. Specifically, our dataset is sourced from four websites: Pexels \cite{pexels}, Artgrid \cite{artgrid}, Airvuz \cite{airvuz}, and Skypixel \cite{skypixel}. We follow Emu3 \cite{wang2024emu3} to split the videos with PySceneDetect \cite{pyscene} to identify content changes and fade-in/out events. Additionally, we remove clips with excessive text using PaddleOCR \cite{paddle}. The detailed composition of our WebVi3D dataset is presented in \cref{tab:data_stat}.

\begin{table}[tb]
\centering
\resizebox{\linewidth}{!}{
\renewcommand{\arraystretch}{1.0}
\setlength{\tabcolsep}{2pt}
\begin{tabular}{c|c|cc|ccc}
\toprule
Website   & Domain    & \# Src. Vids     & Total Hrs.   &\#Fil. Vids   &\#Fil. Clips & Fil. Hrs. \\ \midrule
Pexels   & Open      &  6.18M                & 101.77K          & 0.61M    & 2.65M        & 9.96K             \\
Artgrid  & Open      & 3.94M               &  92.49K            &  0.54M     &  1.10M       & 8.77K      \\
Airvuz   & Drone Shot    &  5.10M                &  94.75K          &  0.54M      &  5.87M   & 8.72K                    \\
    Skypixel & Landscape & 10.27M          &   105.47K            & 0.61M      & 6.37M     & 8.82K                 \\ \midrule
  \textbf{Total}    & Open    & 25.48M         & 394.48K                & \textbf{2.30M  }    & \textbf{15.99M  }          & 36.27K  \\
\bottomrule
\end{tabular}
}
\vspace{-4pt}
\caption{\textbf{WebVi3D Dataset}. Sourced from four open websites, we curate  $\sim$2.30M videos, which are divided into 15.99M clips featuring temporally static scenes with large-range viewpoint.}
\vspace{-17pt}
\label{tab:data_stat}
\end{table}

\vspace{4pt}
However, identifying 3D-aware videos presents a nontrivial challenge.
As most videos are derived from real-world footage, such videos often contains dynamic scenes or  small camera movement. 
To address this, we propose a pipeline that automatically selects relevant, high-quality 3D-aware data (i.e., multi-view images)  by leveraging priors from instance segmentation \cite{maskrcnn}, optical flow \cite{raft}, and pixel tracking \cite{cotracker}. This pipeline comprises four core steps: 

\vspace{2pt}
\noindent \textbf{a) Temporal-Spatial Downsampling.}
To improve data filtering efficiency, we first downsample each video clip both temporally and spatially. The final resolution is set to 480p, and the temporal downsampling rate is set to 2. Note that this downsampling operation is applied  only during data curation, not during model training. 

\vspace{2pt}
\noindent \textbf{b) Semantic-Based Dynamic Recognition.} We employ the instance segmentation model, Mask R-CNN \cite{maskrcnn}, to generate motion masks for potential dynamic objects, such as humans, animals, and sports equipment. A threshold is applied to filter out videos based on the proportion of frames containing these objects, as they are more likely associated with dynamic scenes.  

\vspace{2pt}
\noindent \textbf{c) Flow-Based Dynamic Filtering.} To precisely filter out videos with dynamic regions, we use offline optical flow estimation \cite{raft} to obtain dense matching, which enables us to identify dynamic motion masks in video frames. These masks are then analyzed based on their locations to further determine whether the video contains dynamic content.

\vspace{2pt}
\noindent \textbf{d) Tracking-Based Small Viewpoint Filtering.} The previous three steps yield videos with static scenes.  To further ensure these videos contain multi-view images captured from a larger camera viewpoint, we track the motion trajectory of key points across frames and calculate the radius of the minimum outer tangent circle of the trajectory.
Videos with a small trajectory radius are then filtered out. More details about the data curation pipeline are provided in the Appendix \ref{supp:data}.

\vspace{4pt}
Finally, we curate approximately 320M multi-view images from 15.99M video clips with static content and sufficient multi-view observations (see Fig.\ref{fig:data}b). To validate the effectiveness of our data acquisition method, we randomly select 10,000 video clips for human annotation, of which 8,859 were labeled as 3D-aware, representing 88.6\% of the total. This indicates that our pipeline effectively identifies  3D-aware videos from massive source videos. As the volume of Internet videos continues to grow, this pipeline can continuously acquire more 3D-aware data, allowing for ongoing expansion of our dataset.

\vspace{4pt}
\subsection{Visual Conditional Multi-View Diffusion Model} \label{sec:model}
\vspace{4pt}
\paragraph{Preliminary.}
Diffusion models \cite{sohl2015deep,song2020denoising,ho2020denoising} operate by perturbing the training data $X_0 \sim  q(X_0)$ through a forward diffusion process and learning to reverse it. The forward diffusion process $X_t \sim q_{t|0}(X_t | X_0)$ can be formally represented by $X_{t} = \sqrt{\bar{\alpha}_{t}} \mathbf{X}_0 + \sqrt{1 - \bar{\alpha}_{t}} \boldsymbol{\epsilon}, \quad \boldsymbol{\epsilon} \sim \mathcal{N}(0, \mathbf{I})$, where $\bar{\alpha}_{t}$ is variance schedule used in noise scheduler. In theory, $X_t$ approximates an isotropic Gaussian distribution for sufficiently large timesteps $t$. The training objective is to learn the reverse process. 

\vspace{4pt}
\noindent\textbf{Objective.} We aim for multi-view prediction: generating novel views along specified camera trajectories from a single or sparse input while ensuring consistency with the input appearance. 
The MVD model inherits the generalizability of the 2D diffusion model while capturing cross-view consistency, which naturally aligns with our goal.  Following this line, we present \ours, a pose-free, visual-conditional MVD model trained on Internet videos to enable robust 3D generation, as shown in  Fig.\ref{fig:trainpipe}.

\vspace{4pt}
\noindent\textbf{Challenge.} The main technical challenge lies in learning precise camera control from pose-free videos. 
Previous works commonly incorporate camera parameters for both input and target views into diffusion models to guide multi-view generation from specified viewpoints. However, training these models generally requires expensive 3D data with precise camera pose annotations, which limits scalability. 
To address this,  we explore an alternative approach that conditions on 2D-inductive visual hints to implicitly control camera movement during training, thereby avoiding the need for hard-to-obtain camera trajectories. 

\vspace{4pt}
\noindent\textbf{Formulation.} Formally, we propose training the MVD model conditioned on 2D-inductive visual signals, referred to as \textit{visual-condition}, without incorporating  camera parameters. This task can be formulated as designing a conditional distribution, achieved by a conditional diffusion model that minimizes:
\begin{equation}
\label{eq:loss}
\mathbb{E}_{X_0, Y_0, \epsilon, t} \left[ \left\lVert \epsilon_{\theta}(X_t, Y_0, V, t) - \epsilon \right\rVert_2^2 \right],
\end{equation}
where $X_t$ denotes the noisy latent. $X_0=\left \{ x_{0}^{i}  \right \} _{i=1}^{N}$ represents a multi-view observation of 3D content, formed by sampling one clip from  \textit{\data}  as described in Section \ref{sec:data}, with $N=S+L$ being the number of frames in each clip.  
From $X_0$, $S$ frames are randomly selected as reference views, noted as $Y_0=\left \{ y_{0}^{i}  \right \} _{i=1}^{S}$,  while the remaining $L$ frames are treated as target images, denoted $G=\left \{ g^{i}  \right \} _{i=1}^{L}$. Our approach focuses on constructing the \textit{visual-condition} $V$, which guides the diffusion model to generate plausible 3D content estimates from target viewpoints, ensuring consistency with the appearance of $Y_0$. 

\vspace{-4pt}
\subsubsection{Principle of Visual-Condition}
A desirable \textit{visual-condition} should meet the following criteria: a) it can be constructed without the need for additional 3D annotations, b) it is independent of specific downstream tasks, and c) it offers sufficient generalization to support various task-specific visual conditions, enabling precise control of camera movements.

\vspace{4pt}
Ideally, this \textit{visual-condition} can be derived  from  pixel-space hints within the original videos, implicitly guiding the model to learn camera control.
Moreover, it should be robust enough to handle domain gaps between task-specific visual cues and pixels extracted from video data.
For example, in warping-based generation, warped images often suffer from issues like self-occlusions, artifacts, and distortions, creating a significant gap compared to real video data as shown in Fig.\ref{fig:single-3d} and Fig.\ref{fig:Vt-ablation}. 

\subsubsection{Time-dependent Visual Condition}
Building  on the analysis above, we propose constructing the \textit{visual-condition} by applying masks, noise, and mixture to the input video data.

\vspace{2pt}
\noindent \textbf{Random Masking:} We first corrupt target images $G$ through random irregular masking  to reduce reliance on direct pixel-space visual signals, helping the model partially mitigate the domain gap between task-specific visual cues and video data. Meanwhile, we keep the reference images $Y_{0}$ clean to provide effective appearance signals.

\vspace{4pt}
\noindent \textbf{Time-dependent Noise:}
We further add noise to video data to approximate a Gaussian distribution. For downstream tasks, task-specific visual inputs are similarly noised, aligning their distributions with this Gaussian profile and further bridging the gap between video data and task-specific inputs.
A key challenge lies in determining  the optimal noise level: excessive noise weakens conditional signals, resulting in poor visual quality and inaccurate camera control, whereas insufficient noise preserves too many details from the target images, causing the model to over-rely on visual hints from the video data.

\vspace{4pt}
Previous studies \cite{muller2024multidiff,singh2022conditioning,hou2024training,zhao2024identifying,everaert2024exploiting} have explored modulating noise levels by adding noise to input data. Notably, as pointed out in the previous work \cite{zhao2024identifying}, diffusion models tend to over-rely on the conditional image at larger time steps, leading to signal leakage. Inspired by this\cite{zhao2024identifying}, we introduce time-dependent noise to the corrupted target images. In addition, we develop a function $t'=f(t)$ to regulate signal leakage, preventing excessive noise from completely obscuring visual cues and disabling camera control. Specifically, we define:
\begin{equation} \label{eq:ct}
C_{t} = \sqrt{\bar{\alpha}_{t'}} (1-M)\mathbf{X}_0 + \sqrt{1 - \bar{\alpha}_{t'}} \boldsymbol{\epsilon}. \quad \boldsymbol{\epsilon} \sim \mathcal{N}(0, \mathbf{I}).
\end{equation}

\noindent Here, $f$ is a strictly monotonically increasing function, ensuring $t' < t$, so that $C_t$ contains at least as much information as $X_t$ at earlier timesteps. $\bar{\alpha}_{t'}$ are the variances used in DDIM \cite{song2020denoising}. A detailed explanation  of $f(t)$ can be found in Appendix \ref{supp:ftwt}.

\vspace{4pt}
\noindent \textbf{Time-dependent Mixture:} However, as $t$ decreases, lower  noise levels increase the risk of signal leakage, causing a domain gap between the video data  and task-specific visual condition distributions.  To address this issue, we propose gradually replacing the corrupted data $C_t$ with noisy latent variables $X_t$ as  timestep decreases. This encourages the model to rely more on pixel-space signals from video data at larger time steps, and transition to $X_t$ at smaller timesteps. To achieve this, we further introduce a weighting factor $W_{t} \in [0,1]$, which decreases monotonically with the timestep $t$, to combine $C_t$ and $X_t$. Formally, our final \textit{visual-condition} is defined as:
\begin{equation} \label{eq:mix}
V_t=[W_t\ast C_t+  \left ( 1-W_t \right ) \ast X_t;M],
\end{equation}
\noindent where $M = \left\{  m^{0:S} \cup m^{S+1:N} \right \}$, with $m^{0:S}$ as a zero matrix, keeping the reference images $Y_0$ unmasked, and $m^{S+1:N}$ as random irregular masks applied to the target images $G$. $V_t=\left \{ v_{t}^{i}  \right \} _{i=1}^{N}$ represents  a mixture of $C_t$ and $X_t$, concatenated with masks $M$ along the channel dimension. 
In practice, an additional processing step assigns $v_{t}^{0:S}$ to the reference images $Y_0$ directly, in order to inject the clean information of $Y_0$ into the model, facilitating alignment between the predicted images and the reference images.
Consequently, Eq.\ref{eq:loss} can be reformulated as $\mathbb{E}_{X_0, Y_0, \epsilon, t} \left[ \left\lVert \epsilon_{\theta}(X_t, Y_0, V_t, t) - \epsilon \right\rVert_2^2 \right]$. A more detailed  definition of $W_t$ is provided in Appendix \ref{supp:ftwt}.

\vspace{4pt}
\subsubsection{Model Architecture}
Our model architecture is based on  video diffusion model \cite{svd}. However, we removed the time embedding, as we aim for the model to control the camera movement purely through visual conditions, rather than inferring movement trends based on temporal cues. To further minimize the effect of temporality, we shuffle the frames in each video clip, treating the data as unordered $X_0$. 
Specifically,  we randomly select a subset of frames from a video clip as reference images, with the remaining frames as target images. The number of reference images is randomly selected to accommodate different downstream tasks.  The multi-view diffusion model is optimized by calculating the loss only on the target images, as described in Eq.\ref{eq:loss}. Additional details regarding the model architecture, including the design of self-attention layers, Zero-Initialize, trainable parameters, noise schedule, and cross-attention, can be found in the Appendix \ref{supp:model}.

\subsection{Visual Conditional 3D Generation} \label{sec:3DGen}
\begin{figure*}[tb]
\centering
\includegraphics[width=0.95\linewidth]{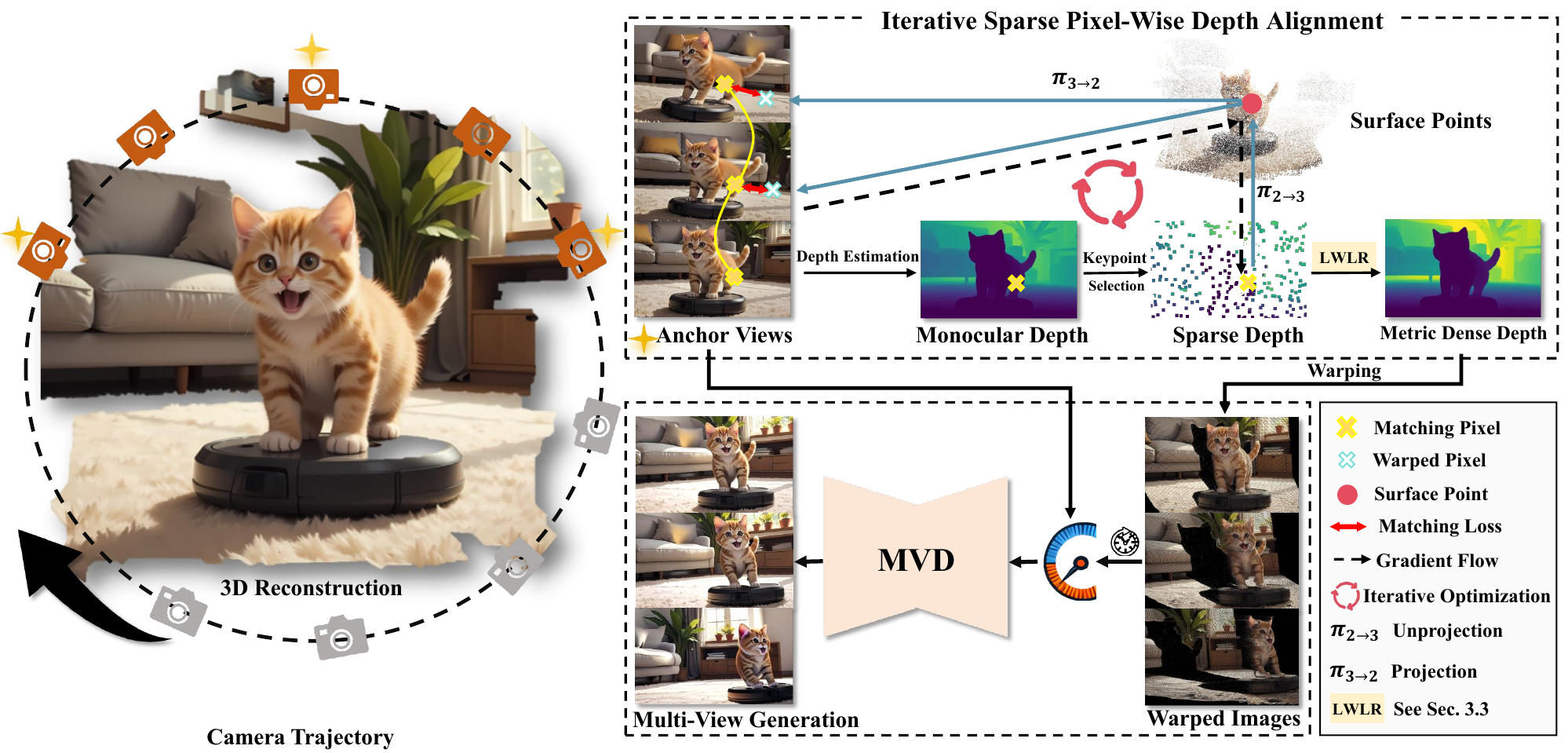}
\caption{\small \textbf{\ours for Multi-View Generation}: From  iteratively generated views ({\color{brown}brown camera}), we randomly select a few anchor views (yellow stars) to guide  the generation of  target views along the {\color[RGB]{122,124,126}\textbf{gray camera trajectory}}. 
Keypoint matching is first performed to establish correspondences between the anchor views.
Next, monocular depth estimation is applied to the latest anchor view, followed by our 
 \textit{Iterative Sparse Pixel-Wise Depth Alignment} to refine the depth and recover a dense map. This dense depth is then used to warp images along the gray camera viewpoints. Subsequently, the warped images and anchor images are combined and processed according to Eq.\ref{eq:ct} and  Eq.\ref{eq:mix}, without random masking, forming the \textit{visual-condition}, which guides  MVD model to produce  3D-consistent target views. Finally, the {\color[RGB]{122,124,126}\textbf{gray camera}} turns to {\color{brown}brown}, guiding multi-view generation in the next iteration.
} 
\vspace{-15pt}
\label{fig:infer}
\end{figure*}

\noindent\textbf{Overview. }
This section demonstrates the application of \ours for domain-free 3D generation, supporting long-sequence novel view  synthesis  with complex camera trajectories. Starting with one or a few input views, we iteratively generate warped images as visual hints, guided by predefined camera poses and estimated global depth \cite{bhat2023zoedepth}. \ours is then utilized  to generate novel views along the predefined camera trajectory, conditioned on the proposed \textit{visual-condition}. This iterative pipeline is illustrated in Fig.\ref{fig:infer}, where the brown cameras represent the already generated views, and the gray cameras indicate the target views we aim to generate.

\vspace{4pt}
\noindent \textbf{Challenge.}
Recent warping-based 3D generation approaches  \cite{chung2023luciddreamer, fridman2024scenescape, lei2023rgbd2} rely on monocular depth or point clouds, and perform global point-cloud alignment to recover the actual geometry for subsequent generations. However, as the reference view often provides a limited scene observation, using offline methods tends to suffer from \textit{scale ambiguity} and \textit{geometric estimation errors}.  Moreover, previous methods often overlook correcting these geometric errors,  leading to distortions and stretching artifacts. These errors accumulate during iterative generation, severely degrading the generation quality. To address this, we propose an iterative  strategy with sparse pixel-wise depth alignment, comprising two core steps: pixel-wise depth scale alignment and global metric depth recovery.

\vspace{4pt}
\noindent \textbf{Pixel-wise Depth Scale Alignment.} We introduce  pixel-wise depth scale alignment using sparse keypoints. This approach performs high-degree-of-freedom independent optimization for all keypoints by leveraging multi-view matching priors from anchor views. Each keypoint independently identifies its multi-view correspondences, allowing for the recovery of both depth scale and surrounding geometry. The corrected scale is then propagated across the entire depth map using 2D distances between keypoints and their neighbors.

% \vspace{4pt}
Specifically, denote $\{T_i\}_{i=0}^{N}$ the predefined camera trajectory. Assuming we have generated $n$ images $\{I_i\}_{i=0}^n$, we now proceed to generate the next $m$ views using the warped image from the last anchor view $I_n$,  referred to as the source view. We first utilize the pre-trained MoGe \cite{wang2024moge} to estimate the affine-invariant depth $\boldsymbol{\hat{D}}_n$ of $I_n$. Inspired by \cite{xu2023frozenrecon}, we perform sparse alignment with $1024$ pairs of matching keypoints $\{\mathbf{m}_n, \mathbf{m}_i\}_k$, obtained by the pre-trianed extractor SuperPoint \cite{detone2018superpoint} and feature matcher LightGlue \cite{lindenberger2023lightglue}. For each matched point, we optimize the corresponding scale $\alpha^k$ and shift $\beta^k$ parameters, where $k \in [0, 1024]$, Our core idea is to recover the depth scaling by minimizing the $L_2$ distance of re-projection between matching points. For each iteration, the warping operation $\Pi_{n\rightarrow i}$ transforms pixels from the source image’s coordinate frame to the target image’s coordinate frame, formulated as: $\Pi_{n \rightarrow i}(\hat{d}_n) =  \hat{d}_n K_{i} T_{i} T_{n}^{-1} K_{n}^{-1}$, where $K_i,  K_n,T_i, T_n$ represent the intrinsic and extrinsic parameters of the source and target frames, respectively. The alignment for each pair is performed using normalized coordinates, ensuring that the warping aligns with the matching prior:
\begin{equation}
    \alpha^{k*},\beta^{k*} = \mathop{argmin} \limits_{\alpha^{k},\beta^{k}} || \hat{d}_n^{k*} K_i T_i T_n^{-1} K_n^{-1} m_n^t - m_i^t||_2^2,
\end{equation}
\noindent where the recovered depth of $k$th pixel is $\hat{d}_n^{k*} = \alpha^{k} \odot \hat{d}_n^k + \beta^{k}$, the $\odot$ is the pixel-wise Hadamard Product. We minimize the matching loss via gradient descent to obtain best scale $ \alpha^{k*}$ and shift parameters $\beta^{k*}$ for each pixel. By performing individual scale recovery and geometry correction, we decouple the depth correlation among different points, achieving accurate single-view reconstruction.

\vspace{4pt}
\noindent \textbf{Global Metric Depth Recovery.} After that, we set these recovered positions as sparse guidance $\hat{d}_n^*$, and introduce Locally Weighted Linear Regression \cite{xu2023frozenrecon} (marked as LWLR in Fig.\ref{fig:infer}) to recover the whole depth map based on the locations between guided points and the other target points. Denote $(u,v)$ represent the 2D positions of the remaining target points, their depth $\boldsymbol{\hat{D}}_n$ can be fitted to the sparse guided points by minimizing the squared locally weighted distance, which is reweighed by the diagonal weight matrix as:
\begin{equation}
   \begin{gathered}
  \boldsymbol{W}_{u,v}=diag(w_1,w_2,...,w_m),  \\
          w_i=\frac1{\sqrt{2\pi}}\exp(-\frac{\mathrm{dist}_i^2}{2b^2}),
    \end{gathered} 
\end{equation}
where $b$ is the bandwidth of Gaussian kernel, and $\mathrm{dist}$ is the Euclidean distance between the guided point and the underestimated target point. Denote $\boldsymbol{X}$ the homogeneous representation of $\boldsymbol{\hat{D}}_n$, the scale map $\boldsymbol{S}_{scale}$ and shift map $\boldsymbol{S}_{shift}$ of target points can be calculated by iterating every location on the whole image, which can be formulated as:
\begin{equation}
    \begin{gathered}
        \min_{\boldsymbol{\beta}_{u, v}} (\hat{d}_n^* -\boldsymbol{X}\boldsymbol{\beta}_{u,v})^\mathsf{T}\boldsymbol{W}_{u,v}(\hat{d}_n^* -\boldsymbol{X}\boldsymbol{\beta}_{u,v})+\lambda\boldsymbol{S}_{shift}^2, \\
    \boldsymbol{\hat{\beta}}_{u,v}=(\boldsymbol{X}^\mathsf{T}\boldsymbol{W}_{u,v}\boldsymbol{X}+\lambda)^{-1}\boldsymbol{X}^\mathsf{T}\boldsymbol{W}_{u,v}\hat{d}_n^*, \\
    \boldsymbol{\beta_{u,v}} = [\boldsymbol{S}_{scale},\boldsymbol{S}_{shift}]_{u,v} ^ \mathsf{T}, \\
    \boldsymbol{D}_n = \hat{d}_n^* \oplus  \boldsymbol{S}_{scale} \odot \boldsymbol{\hat{D}}_n + \boldsymbol{S}_{shift},
\end{gathered}
\end{equation}

\noindent where $\boldsymbol{D}_n$ is the scaled whole depth map, $\oplus$ is the concatenation operator, $\lambda$ is a $l_2$ regularization hyperparameter used for restricting the solution to be simple. Besides, the explicit constraint of the source frame with the target frames allows each novel view to maintain contextual consistency from preceding generations.

\vspace{4pt}
\noindent \textbf{Novel View Generation.}
After obtaining the aligned depth $\boldsymbol{D}_n$, we generate target visual hints through warping as $\hat{I}_j = \Pi_{n \rightarrow j}(\boldsymbol{D}_n)$. The warped images $\{\hat{I}_j\}_{j=n}^{n+m}$ contain  unfilled regions,  as indicated by the binary warping mask $\{M_j\}_{j=n}^{n+m}$, providing strong visual hints for \ours to perform novel view generation. To ensure strong multi-view consistency between the newly generated sequence and the previous content, we randomly select $k$ anchor views $\{I_k\},k \in [1,N]$ from the earlier generated frames to guide subsequent generation. The generation process is formulated as: $I_j = \ours (\hat{I}_j, M_{j}, \{I_0,I_k\})$.
We iteratively perform depth estimation, alignment, warping, and generation  until all predefined multi-view images are obtained. 

\vspace{4pt}
\noindent\textbf{3D Reconstruction.} \label{sec: 3dgs}
We reconstruct the 3D scene using 3D Gaussian Splatting (3DGS) \cite{3dgs}. 
The training objective is to minimize the sum of photometric loss and SSIM loss, consistent with the original 3DGS approach. Additionally, we introduce a perceptual loss (LPIPS \cite{lpips}) to mitigate subtle \textit{inter-frame} discrepancies in multi-view generated images during 3DGS reconstruction. LPIPS emphasizes higher-level semantic consistency between Gaussian-rendered and generated multi-view images, rather than focusing on minor high-frequency differences.
Furthermore, the potential \textit{inner-frame} diversity may lead to inconsistencies with the corresponding camera poses. Following \cite{fan2024instantsplat}, we implement joint pose-Gaussian optimization, treating camera parameters as learnable variables alongside Gaussian attributes, thereby reducing gaps between generated viewpoints and their corresponding camera poses.

%% file: arxiv/4_exp.tex
\section{Experiments}
In \cref{exp:single} and \cref{{exp:sparse}}, we present the single view and sparse views reconstruction with \ours as prior.  Next, we conduct ablation experiments in \cref{exp: ablation} to validate the effectiveness of the proposed modules. Additional implementation details, more results on open-world 3D creation, and further ablation experiments are provided in the Appendix.

\begin{table*}[ht]
\centering
\setlength{\tabcolsep}{10pt}
\resizebox{1\textwidth}{!}{
\begin{tabular}{c|ccccccccc}
\toprule
        \textbf{Methods} & \multicolumn{3}{c}{\textbf{Tanks-and-Temples \cite{knapitsch2017tanks}}} & \multicolumn{3}{c}{\textbf{RealEstate10K \cite{real10k}}} & \multicolumn{3}{c}{\textbf{CO3D \cite{reizenstein2021common}}} \\
 \midrule
                   \textbf{Single View}                         & PSNR $\uparrow$  & SSIM $\uparrow$  & LPIPS $\downarrow$ & PSNR $\uparrow$ & SSIM $\uparrow$ & LPIPS $\downarrow$ & PSNR $\uparrow$ & SSIM $\uparrow$ & LPIPS $\downarrow$ \\
\midrule
                   LucidDreamer \cite{chung2023luciddreamer}            & 13.11 & 0.314 & 0.485 & 15.24 & 0.545 & 0.357 & 13.90 & 0.412 & 0.473 \\
                   ZeroNVS \cite{sargent2023zeronvs}                 & 13.38 & 0.344 & 0.525 & 15.37 & 0.556 & 0.397 & 14.23 & 0.444 & 0.495 \\
                   MotionCtrl \cite{wang2024motionctrl}               & 14.31 & 0.405 & 0.436 & 16.30 & 0.596 & 0.363 & 16.16 & 0.515 & 0.418 \\
                   ViewCrafter \cite{yu2024viewcrafter}             & \cellcolor{orange!20}19.66 & \cellcolor{yellow!20}0.609 & \cellcolor{orange!20}0.238 & \cellcolor{orange!20}21.93 & \cellcolor{yellow!20}0.797 & \cellcolor{orange!20}0.161 & \cellcolor{orange!20}20.17 & \cellcolor{yellow!20}0.664 & \cellcolor{orange!20}0.283 \\
                   ViewCrafter* \cite{yu2024viewcrafter}             & \cellcolor{yellow!20}19.13 & \cellcolor{orange!20}0.616 & \cellcolor{yellow!20}0.255 & \cellcolor{yellow!20}20.49 & \cellcolor{orange!20}0.802 & \cellcolor{yellow!20}0.183 & \cellcolor{yellow!20}19.07 & \cellcolor{orange!20}0.678 & \cellcolor{yellow!20}0.339 \\
                   \textbf{Ours}                     & \cellcolor{red!20}23.76 & \cellcolor{red!20}0.735 &\cellcolor{red!20}0.191 & \cellcolor{red!20}25.36 & \cellcolor{red!20}0.854 & \cellcolor{red!20}0.146 & \cellcolor{red!20}24.28 & \cellcolor{red!20}0.765 &\cellcolor{red!20} 0.251 \\
\midrule
                   \textbf{Sparse Views (3 Views)}    & \multicolumn{3}{c}{\textbf{LLFF \cite{mildenhall2019local}}} & \multicolumn{3}{c}{\textbf{DTU \cite{dtu}}} & \multicolumn{3}{c}{\textbf{MipNeRF-360 \cite{barron2022mipnerf360}}} \\
 % \cmidrule(lr){2-11}
 \midrule
                   Zip-NeRF$^{\dag}$ \cite{zipnerf}            & 17.23 & 0.574 & 0.373 & 9.18  & 0.601 & 0.383 & 12.77 & 0.271 & 0.705 \\
                   MuRF \cite{xu2024murf}                    & 21.34 & 0.722 & 0.245 & \cellcolor{yellow!20}21.31 & \cellcolor{red!20}0.885 & 0.127 & -     & -     & -     \\
                   FSGS  \cite{zhu2025fsgs}                   & 20.31 & 0.652 & 0.288 & 17.34 & 0.818 & 0.169 & -     & -     & -     \\
                   BGGS \cite{han2024binocular}                    & \cellcolor{yellow!20}21.44 & \cellcolor{orange!20}0.751 & \cellcolor{orange!20}0.168 & 20.71 & 0.862 & \cellcolor{orange!20}0.111 & -     & -     & -     \\
                   ZeroNVS$^{\dag}$ \cite{sargent2023zeronvs}             & 15.91 & 0.359 & 0.512 & 16.71 & 0.716 & 0.223 & 14.44 & 0.316 & 0.680 \\
                   DepthSplat \cite{xu2024depthsplat}              & 17.64 & 0.521 & 0.321 & 15.59 & 0.525 & 0.373 & 13.85 & 0.254 & 0.621 \\
                   ReconFusion \cite{wu2024reconfusion}             & 21.34 & 0.724 & 0.203 & 20.74 & \cellcolor{yellow!20}0.875 & 0.124 & \cellcolor{yellow!20}15.50 & \cellcolor{yellow!20}0.358 & \cellcolor{yellow!20}0.585 \\
                   CAT3D \cite{gao2024cat3d}                   & \cellcolor{orange!20}21.58 & \cellcolor{yellow!20}0.731 & \cellcolor{yellow!20}0.181 & \cellcolor{orange!20}22.02 & 0.844 & \cellcolor{yellow!20}0.121 & \cellcolor{orange!20}16.62 & \cellcolor{orange!20}0.377 & \cellcolor{orange!20}0.515 \\
                   \textbf{Ours}                     & \cellcolor{red!20}23.23 & \cellcolor{red!20}0.768 & \cellcolor{red!20}0.135 & \cellcolor{red!20}28.04 & \cellcolor{orange!20}0.884 & \cellcolor{red!20}0.073 & \cellcolor{red!20}17.35 & \cellcolor{red!20}0.442 & \cellcolor{red!20}0.422 \\
\bottomrule
\end{tabular}%
}
 \caption{\textbf{Quantitative Comparison of Single/Sparse Views Generation.} The top rows are results given single view as input, where ViewCrafter$^{*}$ indicates our re-implemented result. The bottom rows are novel view rendering quality given 3 views as input, where Zip-NeRF$^{\dag}$ and ZeroNVS$^{\dag}$ are modified versions with sparse views input as reported in CAT3D.} \label{tab:single-sparse}
 \vspace{-12pt}
\end{table*}

\subsection{Single View to 3D}
\label{exp:single}
\noindent\textbf{Experimental Setting.} \ours supports multi-view generation from a single input view. Following prior work \cite{yu2024viewcrafter}, our evaluation is conducted on the test split of three real-world datasets with various camera trajectories, including Tanks-and-Temples \cite{knapitsch2017tanks}, RealEstate10K \cite{real10k}, CO3D \cite{reizenstein2021common}. 
We follow the approach in ViewCrafter \cite{yu2024viewcrafter} for constructing easy/hard evaluation sets based on different sampling rates applied to the original videos. 
We re-implement ViewCrafter using the  official code released by \cite{yu2024viewcrafter} to validate our easy/hard set splitting, with results shown as ViewCrafter* in \cref{tab:single-sparse}. We conduct comparisons with warping-based baselines, including LucidDreamer \cite{chung2023luciddreamer}, camera-conditional video generation model MotionCtrl \cite{wang2024motionctrl},  warp-image conditional ViewCrafter \cite{yu2024viewcrafter}, and multi-view diffusion model ZeroNVS \cite{sargent2023zeronvs}. We use the same point cloud rasterization as proposed in ViewCrafter \cite{yu2024viewcrafter} instead of depth-based warping to generate visual conditions for fair comparisons. Following \cite{yu2024viewcrafter}, we evaluate only the visual quality of images generated by multi-view diffusion without rendering novel views through 3D reconstruction. We report PSNR, SSIM, and LPIPS \cite{lpips} as evaluation metrics. Among these, PSNR is a traditional pixel-level metric that measures image similarity, which is significantly affected by viewpoint shifts. As such, PSNR reflects the accuracy of viewpoint control provided by our proposed \textit{visual-condition} in multi-view generation.

\vspace{4pt}
\noindent\textbf{Results.} The quantitative comparison results are presented  in the top rows of \cref{tab:single-sparse}. Only average metrics for the easy and hard sets are reported here, detailed values are available in the Appendix \ref{supp:sv3d}. The results for ViewCrafter* are comparable to those reported in its original paper, confirming successful alignment between our method and the baselines. Numerically, our approach  outperforms all baseline methods across all metrics. Specifically, compared to the re-implemented ViewCrafter, our approach achieves a 4.63 dB improvement, demonstrating its capability to generate high-quality novel views. PSNR further demonstrates significant gains, indicating our proposed \textit{visual-condition} enables  precise camera control. Qualitative results are shown in the top rows of \cref{fig:single-3d}. \ours generates high-quality, realistic content within minutes. Despite limited visual cues provided by the warped images, our method produces more reliable and  realistic results with fewer artifacts.

\subsection{Sparse Views to 3D} \label{exp:sparse}
\noindent\textbf{Experimental Setting.} We extend our model to the sparse-view reconstruction task, evaluating it on three datasets: LLFF \cite{mildenhall2019local}, DTU \cite{dtu}, and Mip-NeRF 360 \cite{barron2022mipnerf360}.
We compare our method against several few-shot 3D reconstruction baselines, including optimization-based method MuRF \cite{xu2024murf}, FSGS \cite{zhu2025fsgs}, and BGGS \cite{han2024binocular}; 
diffusion-based methods CAT3D \cite{gao2024cat3d}, ZeroNVS (modified to handle multi-view input) \cite{sargent2023zeronvs}, and ReconFusion \cite{wu2024reconfusion}; as well as the feed-forward method DepthSplat \cite{xu2024depthsplat}. 
Following the evaluation protocols from \cite{niemeyer2022regnerf,zhu2025fsgs,wu2024reconfusion}, we use 3, 6, and 9 views as input. For few-shot reconstruction, dense multi-view images are generated from sparse views, similar to  CAT3D \cite{gao2024cat3d}, and  3DGS reconstruction is performed with pose optimization  to render test views for evaluation. We report PSNR, SSIM, and LPIPS \cite{lpips} to evaluate  novel view synthesis performance.

\vspace{4pt}
\noindent\textbf{Results.} Qualitative and quantitative results are presented in \cref{tab:single-sparse} and \cref{fig:single-3d}, respectively, with additional comparisons for 3, 6, and 9 input views available in Appendix \ref{supp:svs3d}.  The 3DGS model, trained on dense multi-view images generated by \ours, surpassed state-of-the-art reconstruction models in novel view rendering.  
This indicates its ability to provide high-quality, consistent multi-view support for 3D reconstruction without imposing additional constraints.
Compared to ReconFusion \cite{wu2024reconfusion} and CAT3D \cite{gao2024cat3d}, which also leverage  diffusion priors for sparse-view reconstruction, our model exhibits effective scalability. Qualitative  comparisons in Figure \ref{fig:single-3d}  reveal that NVS results produced by \ours exhibit fewer floating artifacts, suggesting its capability to generate more consistent and high-fidelity multi-view images.
\begin{figure}[!ht]
\centering
\small
\includegraphics[width=0.47\textwidth]{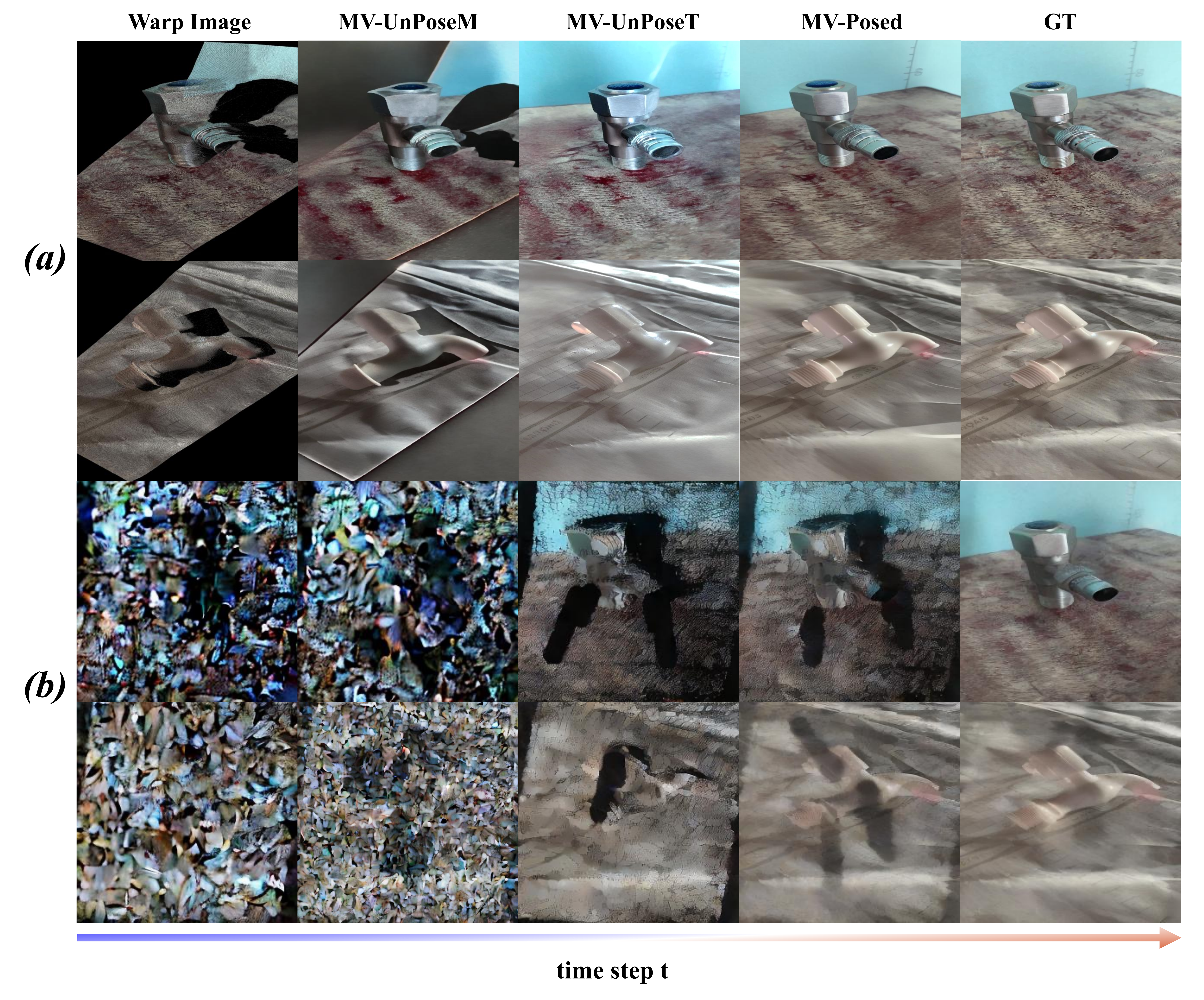}
\vspace{-0.1cm}
\caption{Top: Qualitative ablation of \textit{visual-condition}; Bottom: As timestep decreases, visualize the trend of \textit{visual-condition}.}
\label{fig:Vt-ablation}
\vspace{-0.3cm}
\end{figure}

\begin{figure*}[ht]
\centering
\includegraphics[width=0.85\textwidth]{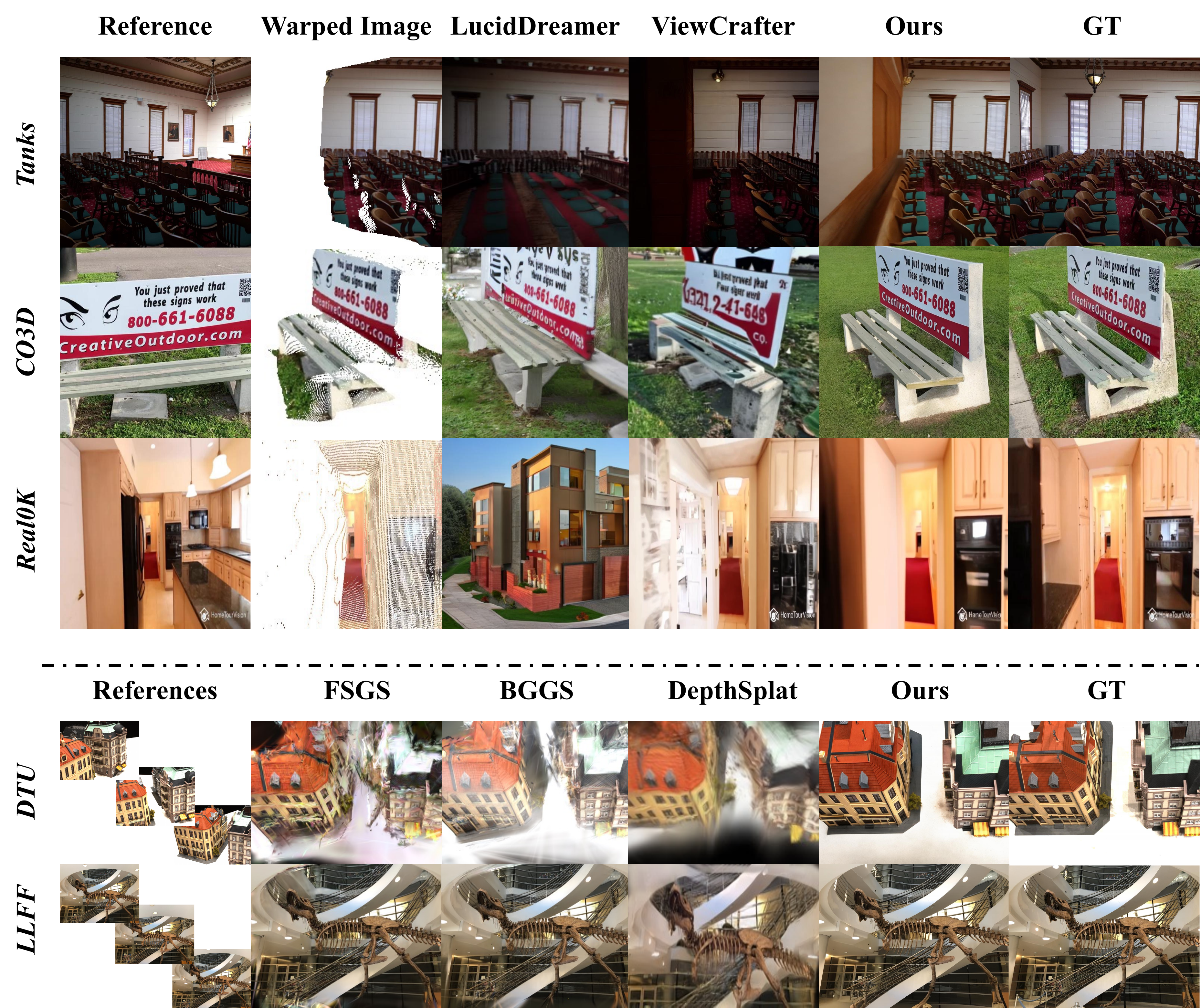}
\vspace{-0.1cm}
\caption{\textbf{Qualitative Comparison of Single/Sparse View Generation.} The top three rows are results with a single view input. The bottom two rows are novel view renderings from 3DGS, where Ours is trained on dense multi-view generation given 3 views as input. Our method outperformed other baselines in capturing high-frequency details, such as text and stairs.}
\vspace{-0.3cm}
\label{fig:single-3d}
\end{figure*}

\subsection{Ablation Study}
\label{exp: ablation}

\noindent\textbf{Scaling up Data.} We investigate the impact of training data by ablating different proportions of our training dataset. The model is trained with 10\%, 20\%, 40\%, 80\%, and 100\% of the training set, and its single-view generation performance is evaluated on RealEstate10K, achieving PSNR values of 19.32, 21.04, 22.57, 24.08, and 25.01, respectively. Additionally, training with unfiltered data results in generated content that often exhibits movement or deformation, leading to a substantial performance drop with a PSNR of 19.55.  We analyze that this degradation likely stems from the lack of stationary and geometrically invariant properties in much of the source video content, which undermines multi-view consistency.  In summary, these findings highlight the critical importance of data quality and diversity for effectively training large-scale MVD models.

\vspace{4pt}
\noindent\textbf{Visual-condition.} Excluding the benefits of data scaling, we investigate  the effectiveness of our \textit{visual-condition} on pose-free data. Previous work \cite{yu2024viewcrafter} has demonstrated that warped images can serve  as a pivot condition to guide the model to generate the target viewpoint. However, due to the reliance on the annotated camera to control the projection and unprojection, warp-based conditions are inherently unscalable. Therefore, we compare the model's ability to control cameras conditioned on pose-free \textit{visual-condition} and conditioned on warped images. Specifically, we extract a subset of  MVImageNet \cite{yu2023mvimgnet} for training and testing. 

\vspace{4pt}
For each multi-view sequence in the training set, we select the point cloud of the first frame and render it into the subsequent 5 camera planes along the camera trajectory, based on the 3D annotations in the dataset. We obtain warped images and form pairs with the ground-truth multi-views to train an MVD model, referred to as MV-Posed. With the same experimental settings (training set, network architecture, batch size and predicted sequence length), we train an additional model without any 3D annotations, except for the modification of warp condition to the time-dependent \textit{visual-condition} $V_t$ described in Sec.\ref{sec:model}, called MV-UnPoseT. Meanwhile, we employ randomly masked multiple views as condition to train the model as an additional baseline, called MV-UnPoseM. 

\vspace{-2pt}
\begin{table}[ht]
\footnotesize
\centering
\setlength{\tabcolsep}{14pt}
\renewcommand{\arraystretch}{1.2} 
\resizebox{0.93\linewidth}{!}{
\begin{tabular}{c|ccc}
\hline
Model &  LPIPS $\downarrow$ & PSNR $\uparrow$ & SSIM $\uparrow$  \\
\midrule
MV-Posed &  0.182 & 26.21 & 0.822  \\
MV-UnPoseM &  0.443 & 16.14 & 0.521  \\
MV-UnPoseT &  0.194  & 25.56 &  0.811  \\
\hline
\end{tabular}
}

\vspace{-4pt}
\caption{\textbf{Ablation Study on Visual-condition}.}
\vspace{-5pt}
\label{tab:abl_Vt}
\end{table}

% \vspace{4pt}
The results are reported in Tab.\ref{tab:abl_Vt} and Fig.\ref{fig:Vt-ablation}, where the performance of MV-Posed and MV-UnPoseT is comparable. In contrast, MV-UnPoseM struggles to handle the gap between the warped image and masked images, in the case of geometric distortion and self-obscuration. These findings indicate that the \textit{visual-condition} offers a viable alternative to 3D-reliant warped conditions. 
Despite a significant domain gap between $V_t$ and warp images as shown in Fig.\ref{fig:Vt-ablation}, our model robustly handles this discrepancy, thanks to the time-dependent nature of the proposed condition.

%% file: arxiv/5_conclusion.tex
\section{Conclusion}
\vspace{-3pt}
We propose a scalable 3D generation framework from the perspective of  dataset scaling, offering  a systematic solution that includes: 1) a new dataset, \data, curated via an automated pipeline, with the potential to evolve with the growing volume of Internet data. 2) a new model, \ours, capable of scalable training without pose annotations, aligning with the concept of `Get 3D by solely Seeing'. 3) a novel \ours-based 3D generation framework that supports long-sequence view generation with complex camera trajectories.
 We show that the 3D priors learned by \ours enable a range of 3D creation applications, including single-view generation, sparse view reconstruction, and 3D editing in open-world scenarios. 
 We believe \ours  provides a new direction to advancing the upper bound of 3D generation through dataset scaling. We hope our efforts will encourage the 3D research community to pay more attention to large-scale unposed data, bypassing the costly 3D data barrier and chasing parity with powerful closed-source 3D solutions.

% \vspace{-9pt}
\paragraph*{Acknowledgments.}
We thank Wenyuan Zhang and Yu-Shen Liu from Tsinghua University, as well as Yance Jiao, Hua Zhou, Liao Zhang, Yaohui Chen, Jinxin Xie, Yiwen Shao, and other colleagues from BAAI, for their valuable support and contributions to the See3D project.

% \clearpage

%% file: arxiv/6_arxiv_suppl.tex
\clearpage
\onecolumn
\tableofcontents
\clearpage

\appendix
\section*{Appendix}
\label{supp}

\vspace{7pt}
\section{ Broader Impact and Limitations} \label{supp:limitation}

\noindent \textbf{Broader Impact}: 
Our model facilitates open-world 3D content creation from large-scale video data, eliminating the need for costly 3D annotations. This can make 3D generation more accessible to industries like gaming, virtual reality, and digital media. By leveraging visual data from the rapidly growing Internet videos, it accelerates 3D creation in real-world applications. However, careful consideration of ethical issues, such as potential misuse in generating misleading or harmful content, is crucial. Ensuring that the data used is curated responsibly to avoid bias and privacy concerns is vital for safe deployment.

\vspace{7pt}
\noindent \textbf{Limitations:} While our model excels at long-sequence generation, it comes with some limitations regarding:
1) Inference Speed: The model requires several minutes for inference, making it challenging for real-time applications. Future work should aim to improve inference speed for real-time generation.
2) Focus on 3D Generation: The current model focuses only on 3D generation, avoiding the modeling of object motion. Future research could extend the model to simultaneously generate 3D and 4D content for dynamic scenes.
3) Model Scalability: While the data scaling approach is effective, the scalability of the model itself has not been explored. Expanding the model's architecture could enhance its capability to handle more complex and diverse 3D content.

\section{ Video Data Curation} \label{supp:data}

Our \data dataset is sourced from Internet videos through an automated four-step data curation pipeline. In this section, we provide further  details on this pipeline process.

\vspace{-10pt}
\paragraph{Step 1: Temporal-Spatial Downsampling.}
To enhance data curation efficiency, we downsample each video both temporally and spatially. Temporally, we retain one frame for every two by downsampling with a factor of two. Spatially, we adjust the downsampling factor according to the original resolution to ensure consistent visual appearance across different video aspect ratios.  The final resolution is standardized to 480p in our experiment.

\vspace{-10pt}
\paragraph{Step 2: Semantic-Based Dynamic Recognition}
We perform content recognition on each frame to identify potential dynamic regions. Following \cite{liu2023robust}, we utilize the off-the-shelf instant segmentation model Mask R-CNN \cite{maskrcnn} to generate coarse motion masks $\mathcal{M}_m$ for potential dynamic objects, including humans, animals, and sports activities. If motion masks are present in more than half of the video frames, the sequence is deemed likely to contain dynamic regions and excluded from further processing.

\vspace{-10pt}
\paragraph{Step 3: Flow-Based Dynamic Filtering}
After filtering out videos with common dynamic objects, we implement a precise strategy to identify and exclude  videos containing  potential dynamic regions, such as drifting water and swaying trees. 
Following \cite{liu2023robust}, we use the pretrained RAFT \cite{raft} to compute the optical flow between consecutive frames. Based on the optical flow, we calculate the Sampson Distance, which measures the distance of each pixel to its corresponding epipolar line. Pixels exceeding a predefined threshold are marked to create a dynamic motion mask $\mathcal{M}_s$. The number of pixels in $\mathcal{M}_s$  serves as an indicator of the likelihood of motion in the current frame.

\vspace{4pt}
However, relying solely on this metric is unreliable, as most data are captured in real shots, where dynamic objects of interest are often concentrated near the center of the imaging plane. These moving regions may not occupy a significant portion of the frame. Therefore, we also consider the spatial location of the dynamic mask and propose a dynamic score $\mathcal{S}$ to evaluate the motion probability for each  frame. Let $H, W$  denote the height and width of an image, respectively. We define the central region as starting at $W'=0.25\times W, H'=0.25\times H$. The proportions of the mask occupying the entire image, $\Theta_i$,  and the central area $\Theta_c$  are calculated as:
\begin{equation}
\Theta_i = \frac{\Sigma_{u,v=0}^{W,H}\mathcal{M}_s(u,v)}{H\times W}, \\
\Theta_c = \frac{\Sigma_{u,v=W', H'}^{W-W',H-H'}\mathcal{M}_s(u,v)}{H/2\times W/2}.
\end{equation}
The dynamic score $\mathcal{S}$ can be formulated as:
\begin{equation}
\mathcal{S}_i =\begin{cases}
     2, & \Theta_i \geq 0.12 \And \Theta_c \geq 0.35 \\
      1.5, &\Theta_i \geq 0.12 \And 0.2 \leq \Theta_c < 0.35 \\
      1, &\Theta_i < 0.12 \And 0.2 \leq \Theta_c < 0.35 \\
      0.5, &\Theta_i < 0.12 \And \Theta_c < 0.2 
\end{cases}.
\end{equation}

This strategy targets the dynamic regions near the image center, enhancing data filtering accuracy. The final dynamic score $\mathcal{S}$ for the entire sequence is calculated as:
\begin{equation}
    \mathcal{S} = \Sigma_{i=0}^N \mathcal{S}_i,
\end{equation}
where $N$ represents the total number of extracted frames. If $\mathcal{S} >= 0.25 \times N$, the sequence is classified as dynamic and subsequently excluded.

\vspace{-8pt}
\paragraph{Step 4: Tracking-Based Small Viewpoint Filtering.}
The previous steps produced videos with static scenes. We require videos that contain multi-view images captured from a wider camera viewpoint. To achieve this, we track the motion trajectory of key points across frames and calculate the radius of the minimum outer tangent circle for each trajectory. Videos with a substantial number of radii below a defined threshold are classified as having small camera trajectories and are excluded. This procedure includes keypoint extraction, trajectory tracking, and circle fitting using RANSAC (Random Sample Consensus) \cite{ransac}.

\vspace{4pt}
\noindent \textit{Keypoint Extraction}. To reduce computational complexity, we downsample the extracted video frames by selecting every fourth frame.  SuperPoint \cite{detone2018superpoint} is then used to extract keypoints $\mathbf{K} \in \mathbb{R}^{N \times 2}$ from the first frame, where $N=100$ represents the number of detected keypoints used to initialize tracking. 

\vspace{4pt}
\noindent \textit{Trajectory Tracking}. Keypoints are tracked across all frames using the pretrained CoTracker \cite{cotracker}, which generates trajectories and  visibility over time as:
\begin{equation}
\mathbf{T}_{\text{pred}}, \mathbf{V}_{\text{pred}} = \text{CoTracker}(\mathbf{I}, \text{queries}=\mathbf{K}).
\end{equation}
Here, $\mathbf{I}$ denotes  the input frames, $\mathbf{T}_{\text{pred}} \in \mathbb{R}^{1 \times T \times N \times 2}$ represents the tracked positions of each keypoint over time, and $\mathbf{V}_{\text{pred}} \in \mathbb{R}^{1 \times T \times N \times 1}$ indicates the visibility of each point. 

\vspace{4pt}
\noindent \textit{Circle Fitting.} For each tracked keypoint, a circle fitting method is applied to its trajectory, selecting only frames where the keypoint is visible $(\mathbf{V}_{\text{pred}} = 1)$. Let $\mathbf{T}_{\text{visible}} \in \mathbb{R}^{M \times 2}$ be the filtered points, where $M$ is the number of visible points. We then use the RANSAC-based circle fitting algorithm on $\mathbf{T}_{\text{visible}}$ to determine the circle’s center $\mathbf{c} = (c_x, c_y)$ and radius  $r$:
\begin{equation}
\mathbf{c}, r = \text{RANSAC}(\mathbf{T}_{\text{visible}}).
\end{equation}
The RANSAC algorithm selects random subsets of three points to define candidate circles, computes the inliers, and optimizes for the circle with the highest inlier count and smallest radius. Finally, we count the number of circles with a radius smaller than a specified threshold,  $r \leq 20$:
\begin{equation}
\text{count} = \sum_{i=1}^{N} \mathbb{I}(r_i \leq 20),
\end{equation}
where $\mathbb{I}$ is the indicator function. The mean radius is also computed to provide an overall measure of circular motion. If the number of small-radius circles exceeds  40 and the average circular motion is less than 5, we classify this video as having small camera trajectories.

\paragraph{User Study.}
To verify the effectiveness of our data curation pipeline, we conducted a user study with a randomly selected set of 10,000 video clips before filtering. We require our users to evaluate videos based on two aspects: \textit{static content} and \textit{large-baseline trajectories}. Only videos meeting both criteria are classified as 3D-aware videos. Among these, 1,163 videos met our criteria for 3D-aware videos, accounting for  11.6\% of the total validation set.
After applying our data screening pipeline, we  randomly selected  10,000 video clips for annotation. In this filtered set, 8,859 videos were identified  as  3D-aware, yielding a ratio of 88.6\%, represents a 77\% improvement compared to the previous set. These results demonstrate the efficacy of our pipeline in filtering 3D-aware videos from large-scale Internet videos.

\begin{figure*}[ht!]
\centering
\includegraphics[width=0.95\textwidth]{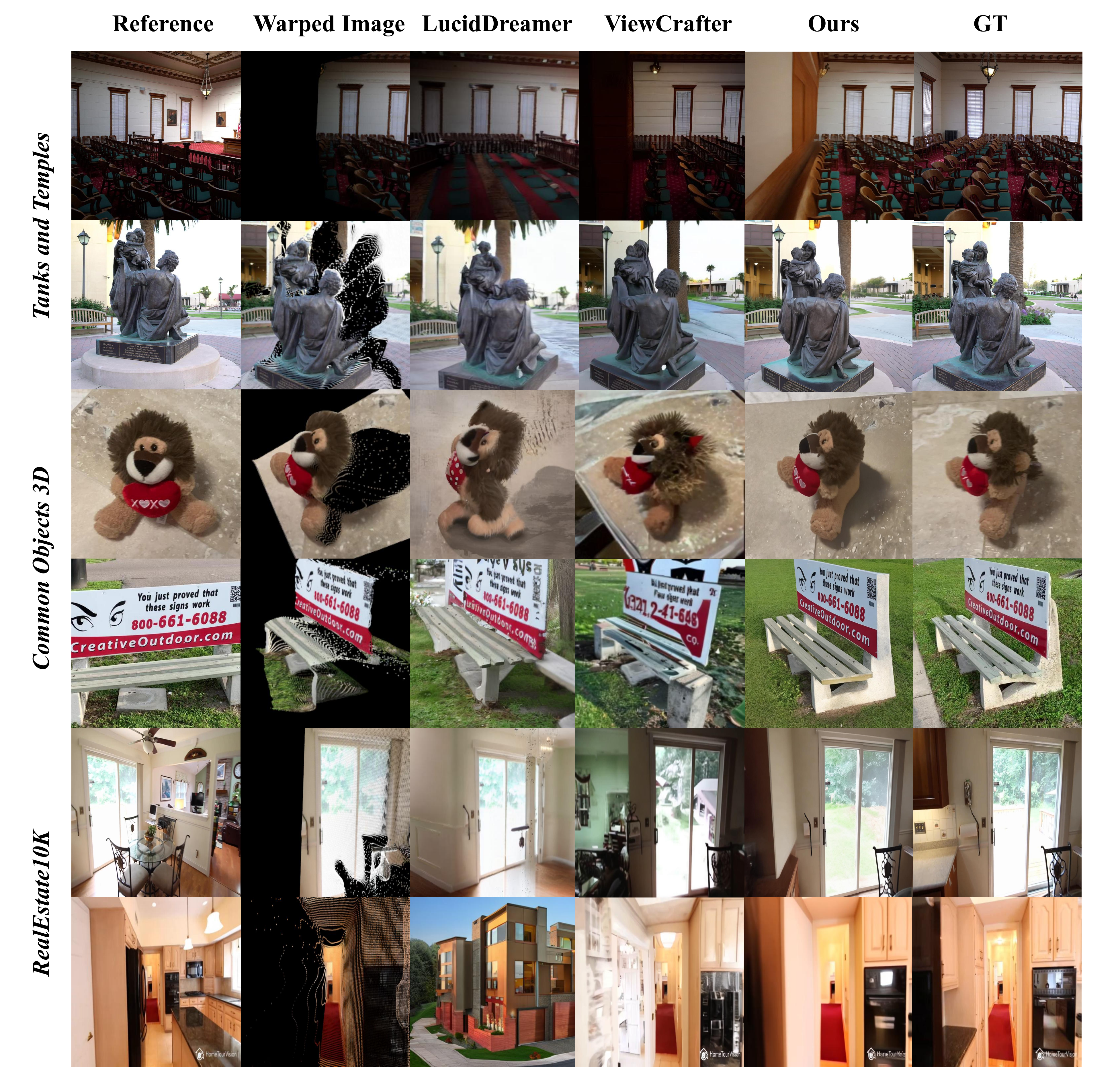}
\caption{\textbf{Single-view to 3D}. Compared with LucidDreamer \cite{chung2023luciddreamer} and ViewCrafter \cite{yu2024viewcrafter}, which are also conditioned on warped images, our model can consistently  generate  high-fidelity views with detailed texture and structural information. }
\label{fig:supp-single}
\end{figure*}

\begin{figure*}[ht]
\centering
\includegraphics[width=0.95\textwidth]{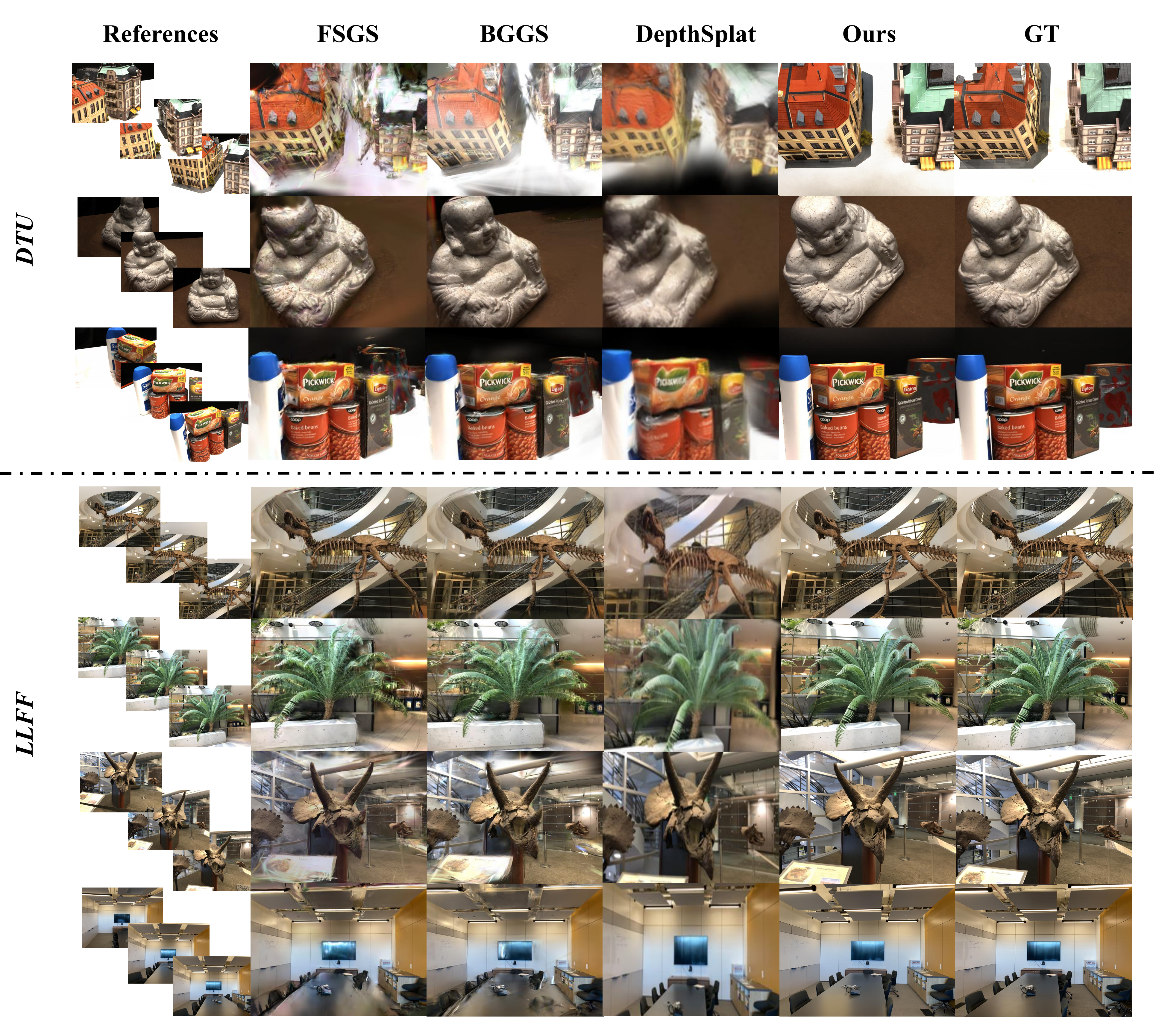}
\caption{\textbf{Sparse-views to 3D}. Given 3 input views, our model generates clear, high-fidelity novel views that closely match the ground truth (GT), without artifacts or blurring. Note that the results from DepthSplat \cite{xu2024depthsplat} are cropped and resized following the same data processing as the official source code.}
\label{fig:supp-sparse}
\end{figure*}

\begin{figure*}[ht]
\centering
\includegraphics[width=0.95\textwidth]{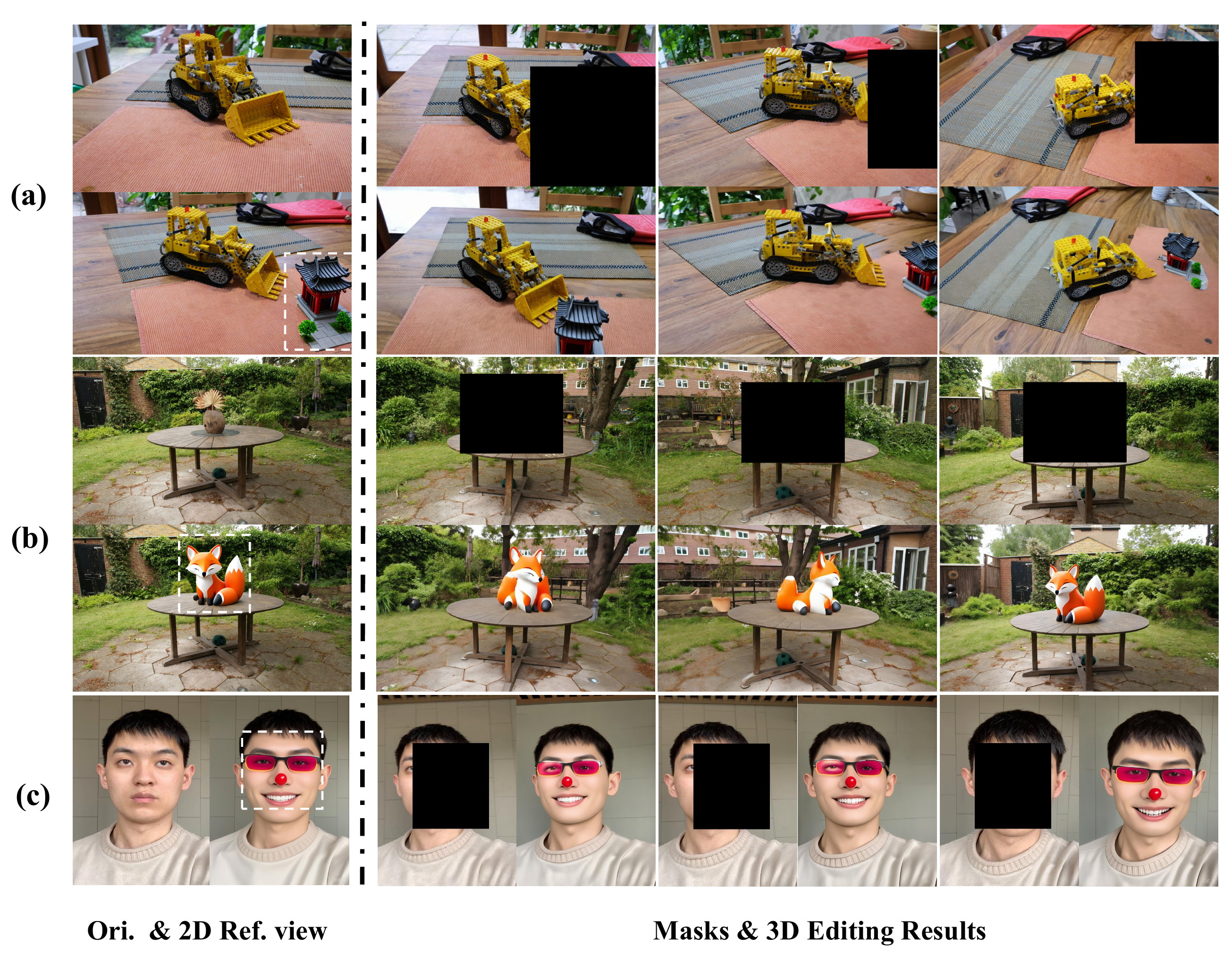}
\caption{
\textbf{Examples of Open-world 3D Editing}. (a) Occlusion-free Editing: An Asian-style attic is added, and novel views are generated realistically. (b) Full Replacement Editing: A vase is replaced with a toy fox, seamlessly integrated  into the scene from various viewpoints. (c) Occluded Editing: Hidden regions in the masked areas are inferred and completed to produce novel views.
}
\vspace{-12pt}
\label{fig:supp-edit}
\end{figure*}

\section{ Technical Implementations} \label{supp:Techimpl}

\subsection{ Model Architecture} \label{supp:model}

The main backbone of \ours model is based on the structure of 2D diffusion models but integrates 3D self-attention to connect the latents of multiple images, as shown in prior work \cite{shi2023mvdream}. Specifically, we adapt the existing 2D self-attention layers of the original 2D diffusion model into 3D self-attention by inflating different views within the self-attention layers. To incorporate visual conditions, we introduce the necessary convolutional kernels and biases using Zero-Initialize \cite{singer2022make}. 
The model is initialized from a pretrained 2D diffusion model \cite{podell2023sdxl} and fine-tuned with all parameters, leveraging FlashAttention for acceleration. In accordance  with prior work \cite{shi2023zero123++}, switching from a scaled-linear noise schedule to a linear schedule is essential for achieving improved global consistency across multiple views.  Additionally, we implement cross-attention between the latents of multiple views and per-token CLIP embeddings of reference images using a linear guidance mechanism \cite{song2022diffusion}. 
For training, we randomly select a subset of frames from a video clip as reference images, with the remaining frames serving as target images. The number of reference images is randomly chosen  to accommodate different downstream tasks.  The multi-view diffusion model is optimized by calculating the loss only on the target images, as outlined in Eq. \ref{eq:loss}.

\subsection{ Training Details} \label{supp:imp}

\noindent \textbf{Brightness Control.} We observe that the \textit{visual-condition} effectively guides camera movement but cannot control brightness changes, posing a significant limitation.
Determining the light source position is particularly challenging with limited observations from single or sparse views.
In our real-world test data, camera movement often causes random highlighting or darkening in some regions of scenes, which has a significant impact on pixel-level metrics like PSNR. This issue highlights a key problem: the inability to control brightness undermines the reliability of pixel-level metrics, as brightness variations affect these metrics more than the actual quality of the generated content. 
To achieve illumination control, 1) we preprocess the training data by converting corrupted images into HSV format, which represents hue, saturation, and brightness. 2) We define a $w \times h$ window and calculate the average brightness difference  within this window between the ground truth image and the corrupted data. 
Using this difference, we apply a scaling factor to the brightness channel of the corrupted data while preserving hue and saturation, before converting the image back to RGB. This ensures brightness adjustment in the \textit{visual-condition} without leaking color or content from the ground truth. 

\vspace{4pt}
During training, we randomly drop this preprocessing with a probability of 0.5, enabling the model to infer lighting changes on its own during inference when brightness control is not required.
In our evaluation experiments, brightness scaling is applied to the unmasked regions of warped images to align with ground truth, reducing the impact of brightness, and thus yielding a higher correlation between the generated content and pixel-level metrics. Meanwhile, keeping hue and saturation unchanged to avoid content or color leakage. Additionally, the model enables user-controlled brightness adjustments for specific regions in multi-view generation by modifying the \textit{visual-condition} as needed.

\noindent \textbf{Training Configuration.} We initialize the \ours model from MVDream \cite{shi2023mvdream} and employ a progressive training strategy.
First,  the model is trained at a resolution of 512 $\times$ 512 with a sequence length of 5. This phase involves 120,000 iterations, using 1 reference view and 4 target views.
Due to the relatively small sequence length,  a larger batch size of 560 is used to enhance stability and accelerate convergence.
Next, the sequence length is increased to 16, and the model is trained for 200,000 iterations with 1 or 3 reference views and 15 or 13 target views, maintaining the resolution of 512 $\times$ 512. In this phase, the batch size is reduced to 228.
Finally, a multi-view super-resolution model is trained using the same network structure. It takes the multi-view predictions from \ours as input and outputs target images with multi-view consistency at a resolution of 1024 $\times$ 1024, using  a batch size of 114. 
In all stages, all parameters of the diffusion model are fine-tuned  with a learning rate of 1e-5. 
% \vspace{4pt}
% \noindent \textbf{Supplementary 3D Data.} 
Additionally, we render some multi-views or extract clips from datasets such as Objaverse \cite{deitke2023objaverse}, CO3D \cite{reizenstein2021common}, RealEstate10k \cite{real10k} , MVImgNet \cite{yu2023mvimgnet}, and DL3DV \cite{ling2024dl3dv} datasets, forming a supplemental 3D dataset with fewer than 0.5M samples, please refer to Section \ref{supp:scad} for details on analysis and ablation. 
During training, this supplemental data is randomly sampled and incorporated into our WebVi3D dataset ($\sim$16M). 
To enhance training efficiency, we utilize FlashAttention \cite{dao2022flashattention} alongside DeepSpeed with ZeRO stage-2 optimizer \cite{rasley2020deepspeed} and bf16 precision.
 We also implement classifier-free guidance (CFG) \cite{ho2022classifier} by randomly dropping visual conditions with a probability of 0.1. 
The \ours model is trained on 114 $\times$ NVIDIA-A100-SXM4-40GB GPUs over approximately 25 days using a progressive training scheme. During inference, a DDIM sampler \cite{song2020denoising} with classifier-free guidance is employed.

\subsection{ Definition of $f(t)$ and $W_{t}$} \label{supp:ftwt}

\paragraph{Definition for $f(t)$.}
In Eq.\ref{eq:ct}, $C_{t}$ is formulated as 
 $C_{t} = \sqrt{\bar{\alpha}_{t'}} (1-M)\mathbf{X}_0 + \sqrt{1 - \bar{\alpha}_{t'}} \boldsymbol{\epsilon}$ , where $\alpha_{t'}$ is a composite function that depends on $\alpha$ and $t'$, with  $t'= f(t)$ and $f(t)=\beta \cdot t$.
 In our experiments, we set the hyper-parameter $\beta=0.2$, which controls the noise level added to $C_{t}$. A larger $\beta$ increases the noise in  $C_{t}$. As $\beta$ approaches 1, $C_{t}$ converges toward a Gaussian distribution,  improving robustness but reducing the correlation between $C_{t}$ and $\mathbf{X}_0$, thereby weakening camera control. 
 Conversely, as $\beta$ approaches 0, the distributions of $C_{t}$ and $\mathbf{X}_0$ become more similar, improving controllability.
However, for downstream tasks, a very small $\beta$ creates  a significant domain gap  between task-specific visual cues and the video data, compromising robustness. Thus, $\beta$  serves as a trade-off parameter, balancing camera control and robustness.

\begin{figure}[h]
\centering
\includegraphics[width=0.5\textwidth]{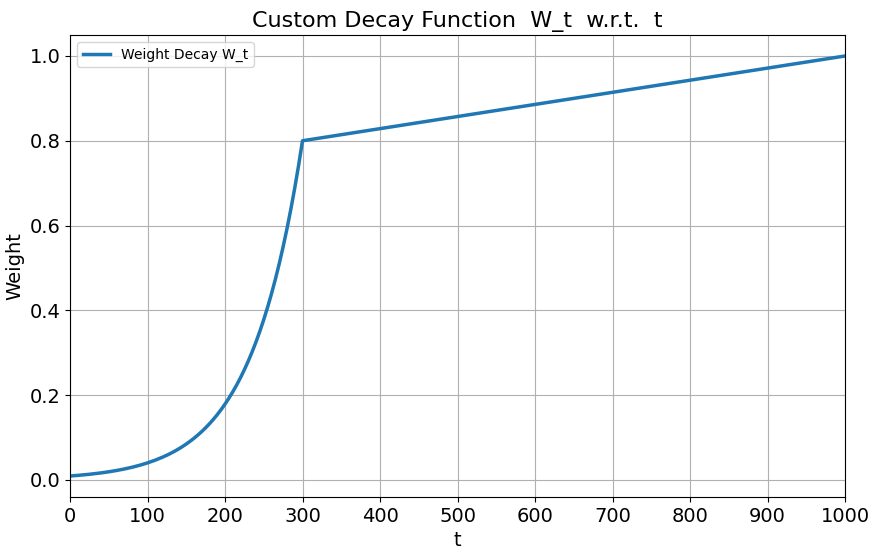}
\caption{\textbf{Piecewise Function $W_t$}, showing linear decay for timesteps $t$ between 300 and 1000, and a monotonically decreasing concave behavior for $t <300$.}
\label{fig:wt}
\end{figure}

\paragraph{Formulation for $W_{t}$.}
Recapping Eq.\ref{eq:mix} from the main manuscript, \( V_t = [W_t \ast C_t + (1 - W_t) \ast X_t; M] \), where \( W_t \) is defined as a piecewise function of \( t \).
\[
W_{t} = 
\begin{cases} 
v_{\text{decay\_end}} \cdot e^{-b \cdot (t_{\text{decay\_end}} - t)}, & \text{if } t < t_{\text{decay\_end}}, \\
1 - \left( 1 - v_{\text{decay\_end}} \right) \cdot \frac{t_{\text{peak}} - t}{t_{\text{peak}} - t_{\text{decay\_end}}}, & \text{if } t \geq t_{\text{decay\_end}},
\end{cases}
\]
\noindent where $t_{\text{peak}}=1000$, $t_{\text{decay\_end}}=300$, $v_{\text{decay\_end}}=0.8$, and $b=0.075$. To ensure that $W_t$  remains within the range \([0, 1]\), it is clamped as: $W_{t} = \text{clamp}(W_{t}, 0, 1)$.  As shown in Figure \ref{fig:wt}, 1) For $t$  between 300 and 1000,  $W_t$  decreases linearly as $t$ decreases; 2) For $t<300$, $W_{t}$ transitions to a monotonically decreasing concave function of $t$. 

\vspace{4pt}
The rationale behind this design is to ensure that when $C_{t}$ has significant noise, it exerts a stronger influence  on $V_{t}$, thus affecting MVD generation. Conversely, as the noise in $C_{t}$ diminishes, $X_{t}$ rapidly  replaces  $C_{t}$, reducing the risk of information leakage from $C_{t}$ and improving the robustness of task-specific visual cues. 
The formulation of $W_{t}$ enables flexible parameter tuning, such as $v_{\text{decay\_end}}$ and $b$, to control its monotonic behavior. Smaller parameter values emphasize the impact of $C_{t}$ on MVD, while larger values prioritize robustness.

\section{ More Experimental Results}

Leveraging the developed web-scale dataset \data, our model supports both object- and scene-level 3D creation tasks, including single-view-to-3D, sparse-view-to-3D, and 3D editing. Additional experimental results for these tasks are presented below.

\subsection{ Single View to 3D} \label{supp:sv3d}

Table \ref{tab:comp_zero} presents a quantitative comparison of zero-shot novel view synthesis performance on the Tanks-and-Temples \cite{knapitsch2017tanks}, RealEstate10K \cite{real10k}, and CO3D \cite{reizenstein2021common} datasets. Our method consistently outperforms all others on both easy and hard sets, achieving the best results in every evaluation metric.
Qualitative results are shown in Figure \ref{fig:supp-single}. Compared to warping-based competitors such as LucidDreamer \cite{chung2023luciddreamer} and ViewCrafter \cite{yu2024viewcrafter}, our approach more effectively captures both geometric structure and texture details, producing more realistic 3D scenes.   These results highlight the robustness and versatility of our method in synthesizing high-quality novel views across diverse and challenging scenarios.

\begin{table*}[ht]
\setlength\tabcolsep{18pt}
\centering
\resizebox{1\textwidth}{!}{%
\small
\begin{tabular}{lcccccc}
\cmidrule[\heavyrulewidth]{1-7}
 \textbf{Dataset}
 & \multicolumn{3}{c}{\textbf{Easy set}} & \multicolumn{3}{c}{\textbf{Hard set}} \\
 \cmidrule(lr){2-4} \cmidrule(lr){5-7}
 Method & LPIPS $\downarrow$ & PSNR $\uparrow$ & SSIM $\uparrow$ & LPIPS $\downarrow$ & PSNR $\uparrow$ & SSIM $\uparrow$ \\ \cmidrule{1-7}

\textbf{Tanks-and-Temples}\\
 LucidDreamer~\cite{chung2023luciddreamer}   & 0.413 & 14.53 & 0.362  & 0.558 & 11.69 & 0.267 \\
  ZeroNVS~\cite{sargent2023zeronvs}  & 0.482 & 14.71 & 0.380  & 0.569 & 12.05 & 0.309 \\
 MotionCtrl~\cite{wang2024motionctrl}  &  0.400 &  15.34 &  0.427  &  0.473 &  13.29 &  0.384   \\
  ViewCrafter & \cellcolor{orange!20}0.194 & \cellcolor{orange!20}21.26 & \cellcolor{orange!20}0.655  & \cellcolor{orange!20}0.283 &  \cellcolor{orange!20}18.07 & \cellcolor{yellow!20}0.563  \\ \cmidrule{1-7}    
  ViewCrafter* & \cellcolor{yellow!20}0.221 & \cellcolor{yellow!20}20.39 & \cellcolor{yellow!20}0.648 & \cellcolor{yellow!20}0.289 & \cellcolor{yellow!20}17.86 & \cellcolor{orange!20}0.584  \\
  Ours & \cellcolor{red!20}0.167 & \cellcolor{red!20}25.01 & \cellcolor{red!20}0.756  & \cellcolor{red!20}0.214 & \cellcolor{red!20}22.52 & \cellcolor{red!20}0.714  \\
  \cmidrule{1-7}    
\textbf{RealEstate10K} \\
 LucidDreamer~\cite{chung2023luciddreamer}     & 0.315 & 16.35 & 0.579  & 0.400 & 14.13 & 0.511   \\
 ZeroNVS~\cite{sargent2023zeronvs}    & 0.364 & 16.50 & 0.577  & 0.431 & 14.24 & 0.535  \\ 
 MotionCtrl~\cite{wang2024motionctrl}  & 0.341 & 16.31 & 0.604  & 0.386 & 16.29 & 0.587  \\
  ViewCrafter &  \cellcolor{orange!20}0.145 &  \cellcolor{orange!20}21.81 & \cellcolor{yellow!20}0.796  &  \cellcolor{orange!20}0.178 &  \cellcolor{orange!20}22.04 &  \cellcolor{orange!20}0.798 
  \\ \cmidrule{1-7}  
  ViewCrafter* & \cellcolor{yellow!20}0.164 & \cellcolor{yellow!20}20.59
 &  \cellcolor{orange!20}0.825 & \cellcolor{yellow!20}0.201 & \cellcolor{yellow!20}20.40 & \cellcolor{yellow!20}0.778  \\
  Ours & \cellcolor{red!20}0.125 & \cellcolor{red!20}26.54 & \cellcolor{red!20}0.872  & \cellcolor{red!20}0.167 & \cellcolor{red!20}24.18 & \cellcolor{red!20}0.837  \\
  \cmidrule{1-7}    
\textbf{CO3D} \\
 LucidDreamer~\cite{chung2023luciddreamer}     & 0.429 & 15.11 & 0.451  & 0.517 & 12.69 & 0.374   \\ 
 ZeroNVS~\cite{sargent2023zeronvs}    & 0.467 & 15.15 & 0.463  & 0.524 &  13.31 &  0.426 \\
 MotionCtrl~\cite{wang2024motionctrl}  & 0.393 & 16.87 & 0.529  & 0.443 & 15.46 & 0.502   \\
 ViewCrafter &  \cellcolor{orange!20}0.243 &  \cellcolor{orange!20}21.38 & \cellcolor{yellow!20}0.687 &  \cellcolor{orange!20}0.324 &  \cellcolor{orange!20}18.96 & \cellcolor{yellow!20}0.641 \\
 \cmidrule[\heavyrulewidth]{1-7}
 ViewCrafter* & \cellcolor{yellow!20}0.331 & \cellcolor{yellow!20}20.12
 &  \cellcolor{orange!20}0.703 & \cellcolor{yellow!20}0.348 & \cellcolor{yellow!20}18.02 &  \cellcolor{orange!20}0.653  \\
  Ours & \cellcolor{red!20}0.225 & \cellcolor{red!20}25.23 & \cellcolor{red!20}0.781  & \cellcolor{red!20}0.276 & \cellcolor{red!20}23.33 & \cellcolor{red!20}0.748  \\
  \cmidrule{1-7}    
\end{tabular}
}
\caption{\textbf{Zero-shot Novel View Synthesis (NVS) on Tanks-and-Temples\cite{knapitsch2017tanks}, RealEstate10K\cite{real10k} and CO3D\cite{reizenstein2021common} dataset.}
}
\label{tab:comp_zero}
\end{table*}

\subsection{ Sparse Views to 3D}  \label{supp:svs3d}
Quantitative comparisons using  3, 6, and 9 input views are presented  in Table \ref{tab:supp-sparse}. The 3DGS model trained on multi-view images generated by \ours outperformed state-of-the-art models in novel view rendering, demonstrating its ability  to provide consistent multi-view support for 3D reconstruction without additional constraints. Qualitative comparisons in Figure \ref{fig:supp-sparse} reveal  fewer floating artifacts in the NVS results, indicating \ours generates higher-quality and more consistent multi-view images.

\begin{table*}[ht]

\centering
\setlength\tabcolsep{20pt}
\resizebox{1\textwidth}{!}{%
\huge
\begin{tabular}{lccccccccc}
\cmidrule[\heavyrulewidth]{1-10}
  \textbf{Dataset}
 & \multicolumn{3}{c}{\textbf{3-view}} & \multicolumn{3}{c}{\textbf{6-view}} & \multicolumn{3}{c}{\textbf{9-view}} \\
  \cmidrule(lr){2-4} \cmidrule(lr){5-7} \cmidrule(lr){8-10}
  Method & PSNR $\uparrow$ & SSIM $\uparrow$ & LPIPS $\downarrow$ & PSNR $\uparrow$ & SSIM $\uparrow$ & LPIPS $\downarrow$ & PSNR $\uparrow$ & SSIM $\uparrow$ & LPIPS $\downarrow$ \\ \cmidrule{1-10}

\textbf{LLFF} \\
  Zip-NeRF$^*$~\cite{zipnerf}     & 17.23 & 0.574 & 0.373 & 20.71 & 0.764 & 0.221 & 23.63 & 0.830 & 0.166 \\ 
  MuRF \cite{xu2024murf} & 21.34 & 0.722 & 0.245 & 23.54 & 0.796 & 0.199 & 24.66 & 0.836 & 0.164 \\
    FSGS \cite{zhu2025fsgs} & 20.31 & 0.652 & 0.288 & 24.20 & 0.811 & 0.173 & 25.32 & \cellcolor{yellow!20}0.856 & 0.136 \\
    BGGS \cite{han2024binocular} & \cellcolor{yellow!20}21.44 & \cellcolor{orange!20}0.751 & \cellcolor{orange!20}0.168 & \cellcolor{orange!20}24.84 & \cellcolor{red!20}0.845 & \cellcolor{orange!20}0.106 & \cellcolor{orange!20}26.17 & \cellcolor{red!20}0.877 & \cellcolor{red!20}0.090 \\
  ZeroNVS$^*$~\cite{sargent2023zeronvs}    & 15.91 & 0.359 & 0.512 & 18.39 & 0.449 & 0.438 & 18.79 & 0.470 & 0.416 \\
  DepthSplat \cite{xu2024depthsplat} & 17.64 & 0.521 & 0.321 & 17.40 & 0.499 & 0.340 & 17.26 & 0.486 & 0.341 \\
  ReconFusion~\cite{wu2024reconfusion}         &  21.34 &  0.724 &  0.203 &  24.25 &  0.815 &  0.152 &  25.21 &  0.848 &  0.134\\
  CAT3D \cite{gao2024cat3d} & \cellcolor{orange!20}21.58 & \cellcolor{yellow!20}0.731 & \cellcolor{yellow!20}0.181 & \cellcolor{yellow!20}24.71 & \cellcolor{orange!20}0.833 & \cellcolor{yellow!20}0.121 & \cellcolor{yellow!20}25.63 & \cellcolor{orange!20}0.860 & \cellcolor{yellow!20}0.107 \\
  Ours & \cellcolor{red!20}23.23 & \cellcolor{red!20}0.768 & \cellcolor{red!20}0.135 & \cellcolor{red!20}25.32 & \cellcolor{yellow!20}0.820 & \cellcolor{red!20}0.104 & \cellcolor{red!20}26.19 & 0.844 & \cellcolor{orange!20}0.098
      \\ \cmidrule{1-10}                
\textbf{DTU} \\
    Zip-NeRF$^*$~\cite{zipnerf}     &  9.18 & 0.601 & 0.383 &  8.84 & 0.589 & 0.370 &  9.23 & 0.592 & 0.364  \\
    MuRF \cite{xu2024murf} & 21.31 & \cellcolor{red!20}0.885 & 0.127 & 23.74 & \cellcolor{red!20}0.921 & \cellcolor{yellow!20}0.095 & 25.28 & \cellcolor{orange!20}0.936 & 0.084 \\
    FSGS \cite{zhu2025fsgs} & 17.34 & 0.818 & 0.169 & 21.55 & 0.880 & 0.127 & 24.33 & 0.911 & 0.106 \\
    BGGS \cite{han2024binocular} & 20.71 & 0.862 & \cellcolor{orange!20}0.111 & \cellcolor{orange!20}24.31 & \cellcolor{orange!20}0.917 & \cellcolor{orange!20}0.073 & \cellcolor{orange!20}26.70 & \cellcolor{red!20}0.947 & \cellcolor{red!20}0.052 \\
 ZeroNVS$^*$~\cite{sargent2023zeronvs}    & 16.71 & 0.716 & 0.223 & 17.70 & 0.737 & 0.205 & 17.92 & 0.745 & 0.200  \\
 DepthSplat \cite{xu2024depthsplat} & 15.59 & 0.525 & 0.373 & 15.061 & 0.523 & 0.406 & 14.87 & 0.478 & 0.451 \\
 ReconFusion~\cite{wu2024reconfusion}         &  \cellcolor{yellow!20}20.74 &  \cellcolor{yellow!20}0.875 &  0.124 &  23.62 & \cellcolor{yellow!20} 0.904 &  0.105 & 24.62 &  0.921 &  0.094\\
  CAT3D \cite{gao2024cat3d} & \cellcolor{orange!20}22.02 & 0.844 & \cellcolor{yellow!20}0.121 & \cellcolor{yellow!20}24.28 & 0.899 & \cellcolor{yellow!20}0.095 & \cellcolor{yellow!20}25.92 & \cellcolor{yellow!20}0.928 & \cellcolor{yellow!20}0.073 \\
  Ours & \cellcolor{red!20}28.04 & \cellcolor{orange!20}0.884 & \cellcolor{red!20}0.073 & \cellcolor{red!20}29.09 & 0.900 & \cellcolor{red!20}0.066 & \cellcolor{red!20}29.99 & 0.911 & \cellcolor{orange!20}0.059
\\ \cmidrule{1-10}                  
\textbf{Mip-NeRF 360} \\
  Zip-NeRF$^*$~\cite{zipnerf}     & 12.77 & 0.271 &  0.705  & 13.61 & 0.284 & 0.663  & 14.30 & 0.312 & 0.633  \\ 
  DepthSplat \cite{xu2024depthsplat}    & 13.85 & 0.254 & 0.621 & 13.82 & 0.260 & 0.636 & 14.48 & 0.288 & 0.602 \\
  ZeroNVS$^*$~\cite{sargent2023zeronvs}    & 14.44 & 0.316 & 0.680 & 15.51 & 0.337 & 0.663 & 15.99 &  0.350 &  0.655 \\
  ReconFusion~\cite{wu2024reconfusion}         &  \cellcolor{yellow!20}15.50 & \cellcolor{yellow!20} 0.358 &  \cellcolor{yellow!20}0.585  & \cellcolor{yellow!20} 16.93 & \cellcolor{yellow!20}0.401 &  \cellcolor{yellow!20}0.544 &  \cellcolor{yellow!20}18.19 &  \cellcolor{yellow!20}0.432 &  \cellcolor{yellow!20}0.511    \\
  CAT3D \cite{gao2024cat3d} & \cellcolor{orange!20}16.62 & \cellcolor{orange!20}0.377 & \cellcolor{orange!20}0.515 & \cellcolor{orange!20}17.72 & \cellcolor{orange!20}0.425 & \cellcolor{orange!20}0.482 & \cellcolor{orange!20}18.67 & \cellcolor{orange!20}0.460 & \cellcolor{orange!20}0.460 \\
  Ours & \cellcolor{red!20}17.35 & \cellcolor{red!20}0.442 & \cellcolor{red!20}0.422 & \cellcolor{red!20}19.03 & \cellcolor{red!20}0.517 & \cellcolor{red!20}0.365 & \cellcolor{red!20}19.89 & \cellcolor{red!20}0.542 & \cellcolor{red!20}0.335 \\
 \cmidrule[\heavyrulewidth]{1-10}
\end{tabular}
}
\caption{\textbf{Quantitative Comparison of Sparse-view 3D Reconstruction}}
% \vspace{-2cm}
\label{tab:supp-sparse}
\end{table*}

\subsection{ 3D Editing} \label{supp:3dit} 

Our model, trained on large-scale videos, naturally supports open-world 3D editing without the need for additional fine-tuning.  Figure \ref{fig:supp-edit} illustrates  three distinct editing scenarios: 
a) \textit{Occlusion-free Editing}. An Asian-style attic is placed next to a toy bulldozer in the original image, which serves as the reference view. Our model generates highly realistic images containing the  Asian-style attic from various new viewpoints.
b) \textit{Full Replacement Editing}. The vase in the original image is completely replaced with a toy fox. Our model generates new scenes from different  viewpoints, seamlessly incorporating the toy fox into the designated area with no residual traces of the vase.
c) \textit{Occluded Editing}. Given an occluded edited image as a reference view, our model can generate multiple novel views within the specified masked regions, inferring and filling in the hidden details of the occluded parts.

\section{ Additional Ablation Studies}

\subsection{ Effectiveness of  Pixel-level Depth Alignment} \label{supp:plda}

We conducted additional ablation experiments to validate the effectiveness of the proposed pixel-level depth alignment. Specifically, we enabled and disabled pixel-level depth alignment when generating novel views through warping and visualized the warped results at a specific generation step. As shown in Figure \ref{fig:supp-pixelalign}, the left image shows  the reference GT image, the middle image  corresponds to warping with pixel-level aligned depth, and the right one depicts  warping without pixel-level aligned depth. The results demonstrate that pixel-level depth alignment not only effectively restores the scale of the depth map but also significantly corrects errors in monocular depth estimation (e.g., the toy's neck and the tabletop). Consequently, integrating our proposed 3D generation pipeline improves generation quality.

\subsection{ Efficacy of Scaling up Data \label{supp:scad}} 
% \vspace{-0.3cm}
\begin{figure}[!ht]
    \centering
    \begin{minipage}{0.45\textwidth}
        \centering
        \setlength{\abovecaptionskip}{2pt}
        \setlength{\belowcaptionskip}{-4pt}
        \includegraphics[width=\textwidth]{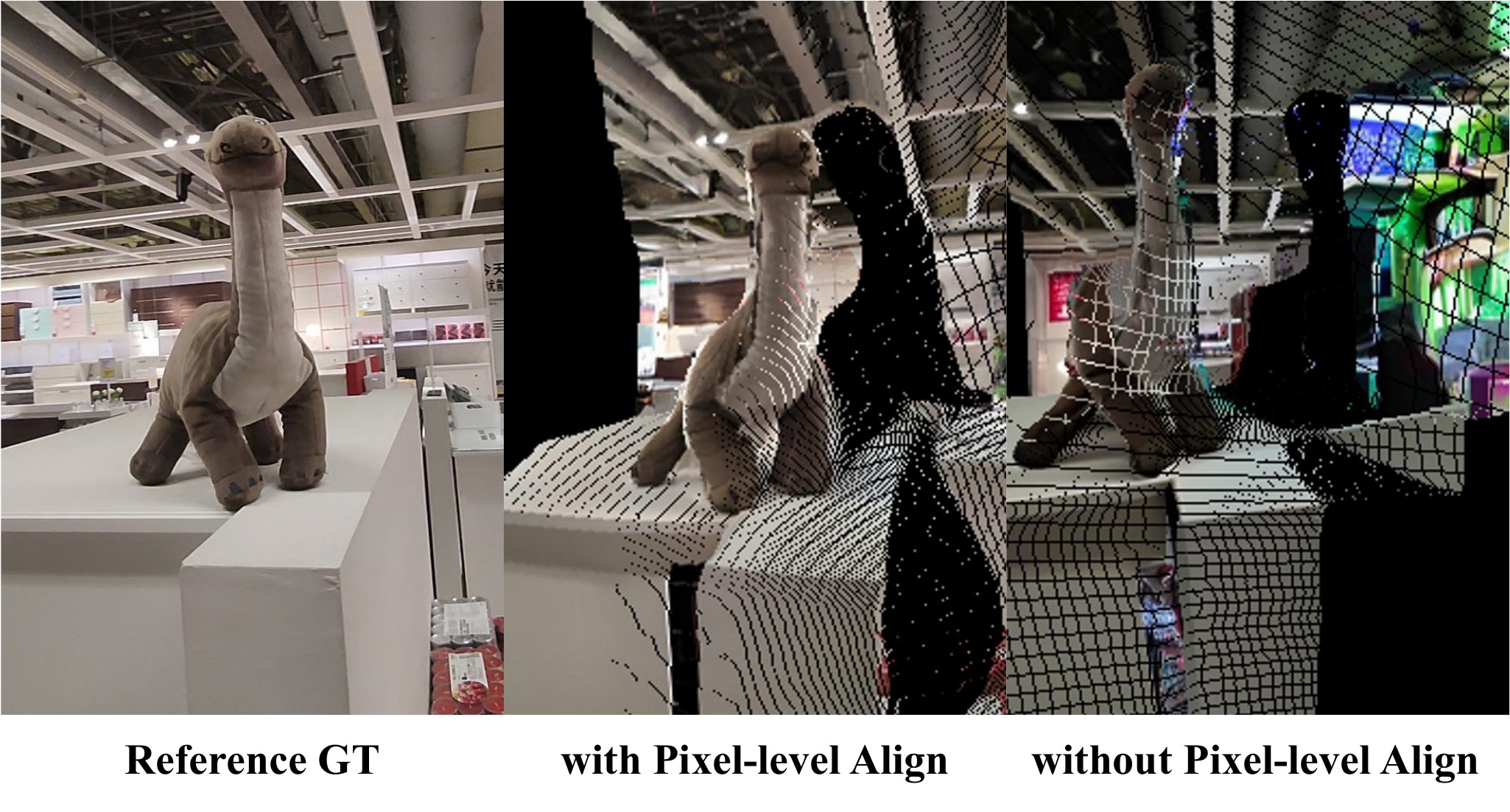} 
        \caption{\textbf{Ablation on Pixel-level Depth Alignment.}}
        \label{fig:supp-pixelalign}
    \end{minipage}
    \hspace{0.02\textwidth}%
    \begin{minipage}{0.52\textwidth}
        \centering
        \small
        \vspace{12pt}
        \setlength{\abovecaptionskip}{20pt}
        \renewcommand{\arraystretch}{1.2}
            \begin{tabular}{cccc}
            \toprule
            Model &  LPIPS $\downarrow$ & PSNR $\uparrow$ & SSIM $\uparrow$  \\
            \midrule
            MV-UnPoseT  &  0.194  & 25.56 &  0.811  \\
            MV-UnPoseT-10\% &  0.187  & 25.95 &  0.817 \\
            MV-UnPoseT-20\% &  0.183  & 26.19 &  0.820  \\
            MV-UnPoseT-60\% &  0.181  & 26.14 &  0.819  \\
            MV-Posed &  0.182  & 26.21 &  0.822  \\
            \bottomrule
            \end{tabular}
        \captionof{table}{\textbf{Ablation on Supplementary 3D Data.}}
        \label{tab:supp_3d}
    \end{minipage}
\end{figure}

In  the main manuscript, we conducted an ablation study on the 3D dataset MVImageNet \cite{yu2023mvimgnet} to evaluate the effectiveness of the proposed \textit{visual-condition}. Table \ref{tab:abl_Vt} shows that:
1) When conditioned on purely masked images, the MV-UnPoseM model performed the worst, struggling with the domain gap issue. 
2) When conditioned on pose-guided warped images, the MV-Posed model achieved the best results, benefiting from pose annotations. 
3) Our MV-UnPoseT model, conditioned on the time-dependent \textit{visual-condition}, demonstrated performance very close to that of the MV-Posed model.
\begin{figure*}[!ht]
\centering
\includegraphics[width=0.85\textwidth]{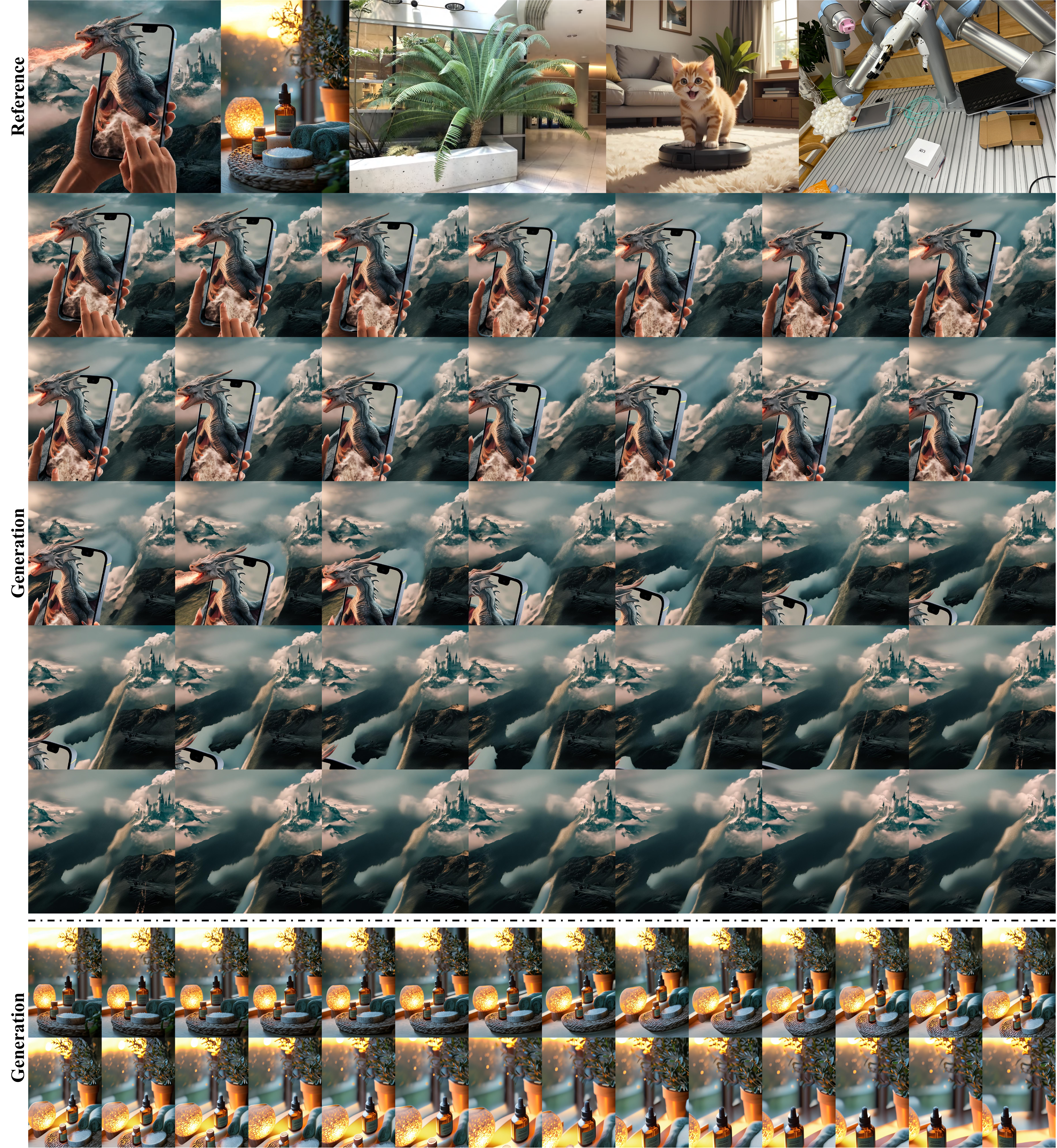}
\caption{
\textbf{Examples of Long-sequence Generation.} High-quality novel views generated along complex camera trajectories, maintaining spatial consistency and visual realism across extended sequences.
}
\label{fig:supp-long1}
\vspace{-0.5cm}
\end{figure*}

\begin{figure*}[ht]
\centering
\includegraphics[width=0.78\textwidth]{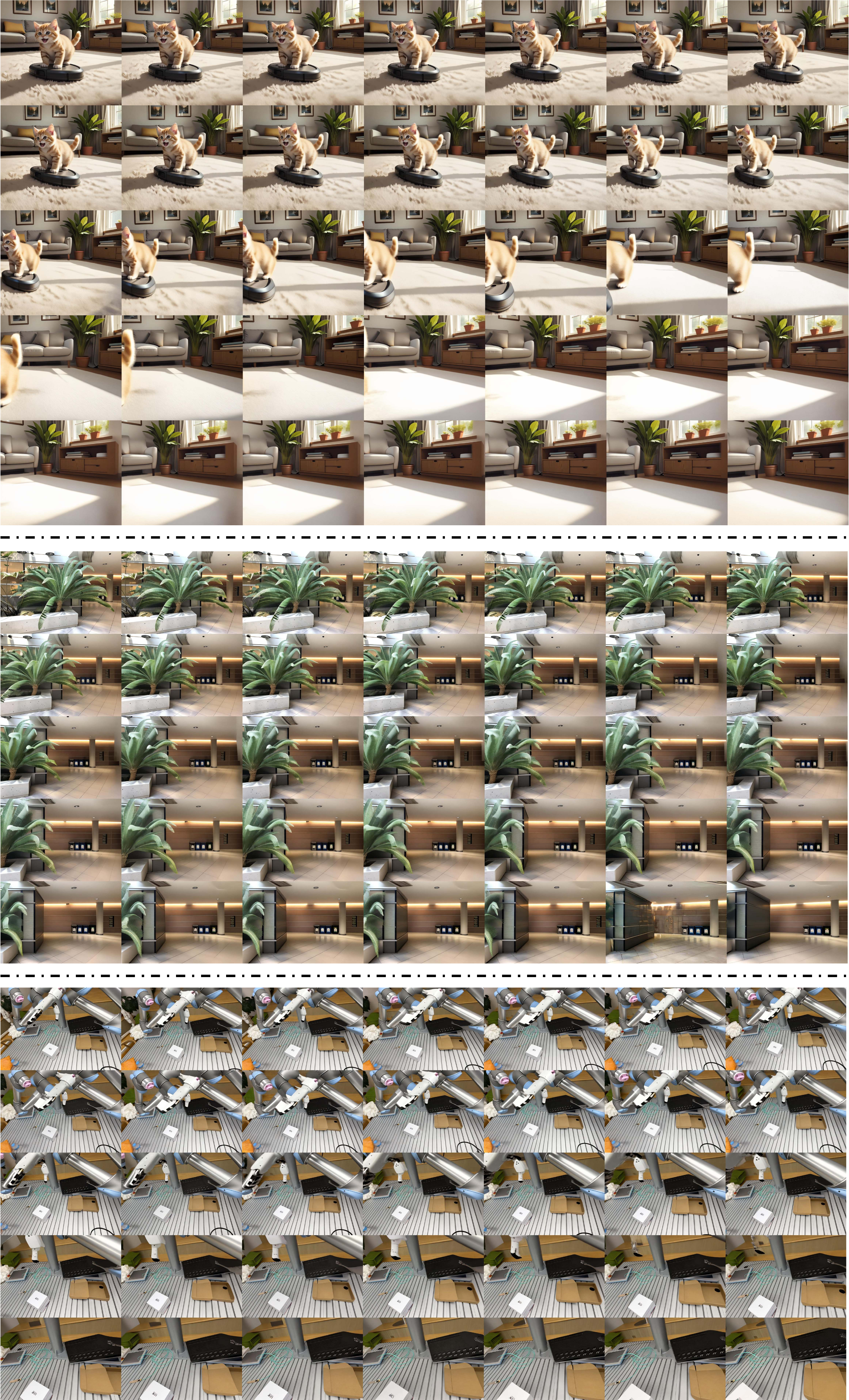}
\caption{
\textbf{More Examples of Long-sequence Generation.}
}
\label{fig:supp-long2}
\end{figure*}

\vspace{4pt}
Intuitively, models trained entirely on 3D data tend to achieve  optimal performance at a specific data scale, establishing an upper bound at that scale.  When the volume of video data matches that of 3D data, models trained on 3D still set the performance ceiling. However, as video data is virtually unlimited, scaling up the dataset can intuitively raise this upper bound.

\vspace{4pt}
Following the same settings in Table \ref{tab:abl_Vt}, we further investigate the impact of supplementing multi-view data with 3D annotations on model performance. We conduct an ablation study using the MV-UnPoseT model, trained on unposed multi-view data with \textit{visual-condition}. In this study, we progressively introduce 3D pose annotations at levels of 10\%, 20\%, 60\%, and 100\% into the training set. When the training data is entirely composed of 3D annotations, the model configuration is equivalent to the MV-Posed model.
The results in Table \ref{tab:supp_3d} indicate that our MV-UnPoseT model, initially trained on unposed data, improves steadily as 3D annotations are introduced. For instance, with only 20\% 3D data (MV-UnPoseT-20\%), the model's performance closely approaches that of the fully 3D-annotated MV-Posed model. This suggests  that even a small amount of 3D data in a largely  unposed multi-view dataset can significantly boost model performance, approaching the  models trained on fully annotated 3D datasets.

\vspace{4pt}
This insight is essential because unposed multi-view data is cost-effective and can be easily collected in large quantities. By incorporating a small volume of high-quality 3D data, we can achieve performance comparable to models trained on large, expensive 3D datasets. Therefore, in our proposed WebVi3D dataset (16M samples), we incorporated a small portion (0.5M samples) of 3D data to optimize model performance.

\vspace{4pt}
\section{ Additional Visualizations} \label{supp:open}
\noindent \textbf{Open-world 3D Generation with Long Sequences.}  We manually configured complex camera trajectories, including rotation, translation, zooming in, zooming out, focus distance adjustments and various random combinations, as shown in Figure \ref{fig:supp-long1} and Figure \ref{fig:supp-long2}. Our model consistently generates high-quality, continuous novel views along these trajectories. Experimental visualizations demonstrate that the model effectively preserves  spatial consistency and visual realism across long sequences. This highlights its robustness in handling intricate camera paths, including rapid transitions and diverse perspectives, making it highly applicable to open-world scenarios.

\clearpage